\documentclass[acmtog]{acmart}
\pdfoutput=1

\renewcommand\footnotetextcopyrightpermission[1]{} 
\usepackage{booktabs} 
\usepackage[nolist, nohyperlinks]{acronym}
\usepackage{amsmath}
\usepackage{bm}  
\usepackage{subcaption}
\usepackage{ltablex}
\usepackage{float}
\usepackage{multirow}

\setcopyright{none}

\newcolumntype{L}[1]{>{\raggedright\let\newline\\\arraybackslash\hspace{0pt}}m{#1}}
\newcolumntype{C}[1]{>{\centering\let\newline\\\arraybackslash\hspace{0pt}}m{#1}}
\newcolumntype{R}[1]{>{\raggedleft\let\newline\\\arraybackslash\hspace{0pt}}m{#1}}

\begin{document}
\title[Active 3D Shape Segmentation]{A Deep Learning Driven Active Framework for Segmentation of Large 3D Shape Collections}

\author{David George}
\affiliation{%
	\department{Department of Computer Science}
	\institution{Swansea University}
    \city{Swansea}
    \country{United Kingdom}}
\email{654214@swansea.ac.uk}

\author{Xianguha Xie}
\affiliation{%
	\department{Department of Computer Science}
	\institution{Swansea University}
    \city{Swansea}
    \country{United Kingdom}}
\email{x.xie@swansea.ac.uk}

\author{Yu-Kun Lai}
\affiliation{%
	\department{School of Computer Science and Informatics}
	\institution{Cardiff University}
    \city{Cardiff}
    \country{United Kingdom}}
\email{laiy4@cardiff.ac.uk}

\author{Gary KL Tam}
\affiliation{%
	\department{Department of Computer Science}
	\institution{Swansea University}
    \city{Swansea}
    \country{United Kingdom}}
\email{k.l.tam@swansea.ac.uk}

\renewcommand\shortauthors{George, D. et al}

\begin{abstract}

High-level shape understanding and technique evaluation on large repositories of 3D shapes often benefit from additional information known about the shapes. One example of such information is the semantic segmentation of a shape into functional or meaningful parts. Generating accurate segmentations with meaningful segment boundaries is, however, a costly process, typically requiring large amounts of user time to achieve high quality results. In this paper we present an active learning framework for large dataset segmentation, which iteratively provides the user with new predictions by training new models based on already segmented shapes. Our proposed pipeline consists of three novel components. First, we a propose a fast and relatively accurate feature-based deep learning model to provide dataset-wide segmentation predictions. Second, we propose an information theory measure to estimate the prediction quality and for ordering subsequent fast and meaningful shape selection. Our experiments show that such suggestive ordering helps reduce users time and effort, produce high quality predictions, and construct a model that generalizes well. Finally, we provide effective segmentation refinement features to help the user quickly correct any incorrect predictions. We show that our framework is more accurate and in general more efficient than state-of-the-art, for massive dataset segmentation with while also providing consistent segment boundaries.
\end{abstract}

%
%



\maketitle
\thispagestyle{empty}

\begin{acronym}
	\acro{ReLU}{Rectified Linear Unit}
	\acro{conv}{convolution}
	\acro{NN}{Neural Network}
	\acrodefplural{NN}[NNs]{Neural Networks}
	\acro{AE}{Autoencoder}
	\acrodefplural{AE}[AEs]{Autoencoders}
	\acro{RF}{Random Forest}
	\acrodefplural{RF}[RFs]{Random Forests}
	\acro{CNN}{Convolutional Neural Network}
	\acrodefplural{CNN}[CNNs]{Convolutional Neural Networks}
	\acro{PCA}{Principal Component Analysis}
	
	\acro{GC}{Gaussian curvature}
	\acro{CF}{Conformal Factor}
	\acro{PC}{Principal Curvature}
	\acro{PCA}{Principal Component Analysis}
	\acro{SDF}{Shape Diameter Function}
	\acro{DMS}{distance from medial surface}
	\acro{AGD}{Average Geodesic Distance}
	\acro{SC}{Shape Context}
	\acro{SI}{Spin Images}
	\acro{HKS}{Heat Kernel Signature}
	\acro{SIHKS}{Scale Invariant \ac{HKS}}
	\acro{LFD}{Light-Field Descriptor}
	\acro{HOG}{Histogram of Oriented Gradients}

	\acro{PSB}{Princeton Segmentation Benchmark}
	
\end{acronym}


\section{Introduction}


In recent years, there has been an increase in the availability of large shape collections. These large and diverse datasets are an invaluable resource for many shape analysis techniques. These dataset are especially necessary for techniques that are driven by a deep learning model as they require large quantities of good and diverse training data \cite{sun2017}. With the efforts from the community, there are many datasets \cite{shilane2004} made available, some even with hundreds of thousands of shapes \cite{wu2015, chang2015}. Recent work has found that analysis tools become more effective when they have access to high quality semantic segmentations \cite{lin2014}. While segmented datasets exist, they either consist of a few hundred shapes \cite{chen2009,sidi2011} or have poor segmentation boundaries \cite{yi2016b} (we show more details on this later). Using these for any shape analysis technique will lead to unreliable results as models are trained on inconsistent data \cite{george2018}.

Segmented datasets have already been shown incredibly useful for many applications, including shape matching \cite{kleiman2015}, retrieval \cite{shapira2010} and modeling \cite{chen2015b}. Shape segmentation techniques often benefit the most from such fully labeled datasets. Supervised techniques require ground truth labels to train segmentation classifiers \cite{kalogerakis2010}, and both supervised and unsupervised techniques need ground truth labels to evaluate their methods \cite{sidi2011}. While existing works have shown good efforts and results \cite{guo2015, qi2017, yi2016}, clear ground truth inconsistencies still exist \cite{george2018}. This means both existing and new techniques could perform better with higher quality ground truth segmentations.

Generating high quality segmentations for shape datasets is a time consuming and interaction heavy task. Smaller datasets, with only small number of inconsistencies or errors may be manageable through manual effort \cite{chen2009,sidi2011}. Massive datasets would take a great amount of user effort however \cite{chang2015}. Further, these massive datasets typically consist of non-manifold (multiple components, holes, zero thickness, etc.) and low-resolution shapes. These shapes are very difficult to process in segmentation pipelines. Recent work try to project them to point clouds \cite{yi2016b, qi2017}, or further to KD-connected point clouds \cite{klokov2017}. While these are viable techniques, and have been shown to work, there may be information loss when using point clouds e.g., connectivity and topology of the shape. Without these, certain reliable features are much harder to compute or are inaccurate when computed (e.g. \ac{SDF} \cite{shapira2008}, Geodesic Distance). Although connectivity can be re-established (e.g. through K Nearest Neighbor, assuming the resolution of the point cloud is high enough), thin regions of the shape could be wrongly connected, leading to undesirable connections. For this reason, in our proposed pipeline, we largely focus on input meshes. We further show that by re-meshing these non-manifold 3D models into manifold meshes, our technique can handle very large dataset very well.

Previous work that generate ground truth segmentations for large dataset typically focus on active learning approach, where a user has some control over the system and influences the decisions in some way. \cite{wang2012} first used an unsupervised co-segmentation algorithm, where the user interactively selects pairs of parts between shapes to connect or disconnect. Recently, \cite{yi2016b} used a supervised algorithm to label a single part at a time. Users are asked to paint two 2D views of a 3D shape. A learning model is trained based on the painted regions and similar shapes (according to global shape descriptors) are evaluated on that model. However, these techniques can only provide a coarse segmentation and output segmentations may have errors. Further \cite{yi2016b} requires one part to be labeled at a time, so datasets with high numbers of parts will take longer and more iterations to label. Here, we developed an active framework which allows full shape segmentation of a shape dataset, to ensure good segmentation quality and it scales well to the number of parts in the dataset.

One of the challenges when developing an active framework for segmentation is minimizing user interactions, while maximizing segmentation quality. To help maximize quality, we utilize a deep learning model for segmentation predictions. In general, deep learning models can take a long time to train, and typically require a large amount of training data. To resolve these, we propose to use a small \ac{CNN}, using two 2D histogram features as input. The features have been shown useful in previous work \cite{kalogerakis2010,guo2015} and fit the \ac{CNN} paradigm as 2D histograms are like images. Our architecture allows for quick model training and we also adopt an ensemble based learning scheme \cite{huang2017} to help generalize with reduced available training data. In our experiments we compare to other feature based \ac{CNN} techniques. We show that our model can perform better than fast existing techniques and comparably with the state-of-the-art.


Another challenge of an active learning framework is the exploration and analysis of model predicted results. It often takes a long time for users to choose the next 3D model to segment, and there are no ground truth data to compare the predictions for ranking. To approach this challenge, we propose to use entropy, a measure of uncertainty, to define a ranking measure without needing ground truth segmentations. The ranking measure provides meaningful ordering of the predicted segment labels in an interactive tabular view. This allows users to see which shapes the deep learning model segmented well or struggled with. Our experiments show that by selecting poorly segmented 3D models, with respect to the ranking measure, it reduces both time and interactions required to segment the whole dataset.

Finally, another problem we observed in existing active framework (e.g. \cite{yi2016b}) is that they do not allow quick boundary refinement. When there are slight errors on the output segmentation, users will likely discard the results, leading to extra manual effort and longer interaction time. With this observation, we propose a segmentation refinement algorithm that takes the current segmentation and information about the shape (e.g. angle and thickness) to refine the segmentation boundaries. This process can be used iteratively. This algorithm can quickly provide high quality segmentations while greatly reducing interaction and time required to refine a shape.

Our proposed framework has been demonstrated to work well on public dataset (including PSB, COSEG), and also on re-meshed dataset from ShapeNet, which contains hundreds of thousands of shapes.

{\textbf{Contributions}} To summarize, the key contribution of this work is a new active learning framework for providing a full segmentation to large sets of 3D shapes. The focus is to maintain accurate and meaningful segment boundaries, while keeping human effort and time at a minimum. There are also several novelties:
\begin{itemize}
	\item First, we show and evaluate a novel deep learning pipeline for shape segmentation which is relatively fast and accurate.
	\item Second, we provide an information-theoretical metric for ranking the performance of shape segmentation algorithms when ground truth data is not available. Our experiment shows that the ordering can help reduces total segmentation efforts and time.
	\item Third, we propose an useful technique for segmentation refinement, which takes into account the segmentation boundaries and thickness of shapes. Our experiment shows that it can help users to quickly improve segmentation boundaries, reducing users efforts and time.
	\item Finally, we provide new and more accurate ground truth segmentations for existing datasets, including massive datasets.
\end{itemize}

In the following, Section~\ref{sec:related_work} discusses the existing work for segmentation, feature extraction and entropy in geometry processing. In Section~\ref{sec:framework_overview}, we briefly overview our active learning framework. Section~\ref{sec:methodology} discuss the details of the three novel subsystem. We further discuss our framework interface and flow in Section~\ref{sec:interface} before outlining our experiments and showing their results in Section~\ref{sec:results}. Finally, in Section~\ref{sec:conclusion} we conclude and discuss possible future work.

\begin{figure*}[!t]
	\centering
	\includegraphics[width=\textwidth]{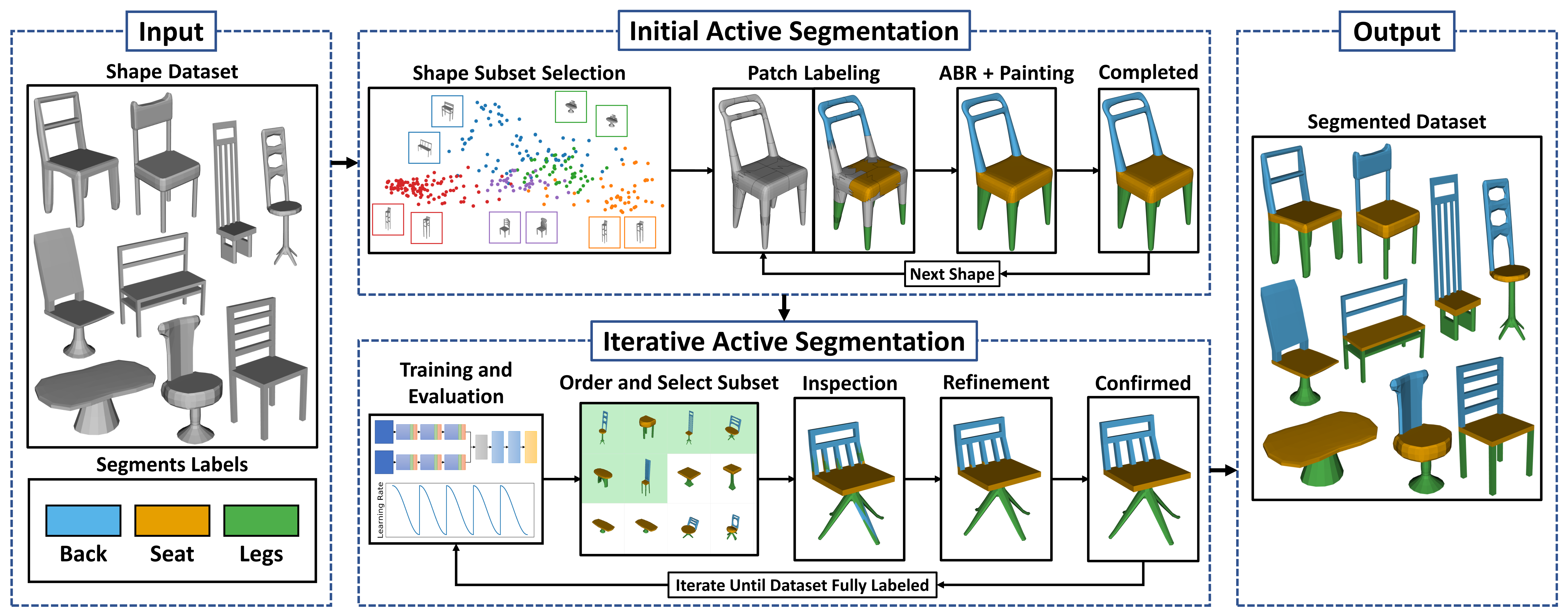}
	\caption{Pipelines for our proposed active framework. For details see: Sections~\ref{sec:input} and \ref{sec:features} (Input Dataset), Section~\ref{sec:lfd} (Shape Subset Selection), Section~\ref{sec:manual-seg} (Patch Labeling, Painting), Section~\ref{sec:boundary-smoother} (Automatic Boundary Refinement (ABR)), Section~\ref{sec:deep-learning} (Training and Evaluation), Section~\ref{sec:table-ordering} (Order and Select Subset), Section~\ref{sec:manual-seg} (Inspection and Refinement).}
	\label{fig:pipeline}
\end{figure*}

\section{Related Work}
\label{sec:related_work}


This work relates to several research areas. We summarize the literature with respect to Shape features, Shape Segmentation, Active learning in image analysis, Active learning in shape analysis, and use of entropy in graphics processing.

\textbf{Shape Features and their Uses} Many of the existing work in shape segmentation are driven by features. These can be defined per face, per vertex, per patch (a cluster of faces), or even per shape.
These features are designed for different purposes, and many have been successfully applied in mesh segmentation.
Per-face features include, \ac{SDF} \cite{shapira2008} which estimates of the thickness of a shape at a given face, \ac{CF} \cite{ben2008} which computes a position invariant representation of the curvature of non-rigid shapes and \ac{SI} which captures the surface information around a face using a 2D histogram. Recent work has also adopted image based features to the 3D domain. One notable example is \ac{SC} \cite{belongie2002} which is a 3D shape descriptor to encode both curvature and geodesic distance distributions in a 2D histogram \cite{kalogerakis2010}. However, there are limitations to how useful a feature can be on certain shapes. Examples are that \ac{CF} is susceptible on shapes with sharp curvature \cite{george2018}, \ac{SDF} can fail if the shape has holes and geodesic distance will fail if the shape has multiple components. Therefore feature selection for a new technique is very important, as it can greatly impact the accuracy and speed. For these reasons, we opted to use two features for this work, \ac{SC} and \ac{SI}. Recent work has shown both features can be very useful in shape segmentation \cite{kalogerakis2010,xie2015,george2018}. Further they are both 2D histograms, so can be generated at any scale (number of bins) and \acp{CNN} should work well to extract useful information.

\textbf{Unsupervised and Supervised Shape Segmentation} 
The goal of a shape segmentation algorithm is to partition a single shape into meaningful parts \cite{shlafman2002,shapira2008}. These algorithms typically used a feature which drives the partitioning (see Features section), though other work also used different strategies like fitting of primitive shapes \cite{attene2006}. Recently, unsupervised techniques looked into co-analysis of a set of shapes, using information consistent across the set to improve the final segmentation \cite{sidi2011,hu2012,meng2013,wu2013,shu2016}. However, these methods struggle with largely varying datasets, especially those with low number of shapes per set \cite{george2018}. Further, the segmentation of parts not only relate to the shape geometry, but the functionality. All these challenges have led to the recent interests in supervised segmentation techniques.


Supervised segmentation techniques rely on prior knowledge in order to train a model. Typically these methods use large pools of shape features as input and classify them according to segment labels \cite{kalogerakis2010}.
Subsequent techniques further improve in different ways, such as ranking features to find segment boundaries \cite{benhabiles2011}, training an extreme learning machine \cite{xie2014, xie2015} to classify the labels. However, similar to unsupervised work, these techniques can struggle when datasets are very diverse.
To combat this, work using \acp{CNN} was proposed \cite{guo2015}. This work arranges a pool of features as an image, and used image based convolution network to predict face labels. However, the simple arrangement lead to unnecessary inference of relationships between features with no correlation, and \cite{george2018} reduce such inference using 1D convolutions, leading to better results. Recently, several techniques have shown new and interesting shape segmentation methods such as point clouds segmentation \cite{qi2016, qi2017}, kd-tree point cloud segmentation \cite{klokov2017}, projecting image segmentations to shapes \cite{evangelos2017}, hierarchal segmentations \cite{yi2017} and graph \acp{CNN} \cite{yi2016}.

With the recent surge of new segmentation papers, each focusing on larger datasets, the need for high quality ground truth labels is very high. 
However, currently available ground truths for widely used segmentation datasets have been shows to contain inconsistent and poor labels for certain shapes within the dataset \cite{george2018}. This can impact the training performance by introducing inconsistent labels for similar samples. It can also impact evaluation, as inconsistencies incorrectly display the performance of a model. Due to this, we emphasize providing accurate, high quality segmentations in this work.

\textbf{Active Image Analysis} Active learning image analysis systems have been widely explored to leverage the human user input to explore large dataset.
They focus on using user input to aid the classifiers by annotation (painting, strokes) or drawing bounding boxes.  This has the advantage that, the user can see what data the classifier is struggling with and incrementally provide new training data to alleviate this problem, making the classifier more generalized and accurate \cite{vijayanarasimhan2009,branson2011,vezhnevets2012,branson2014}. We utilize this functionality in 3D segmentation by allowing the user to incrementally tune the output labels of our model to make it generalize better, while also incorporating a sorting method to easily rank the outputs of the mode, and aid the user in selecting shapes to tune.

\textbf{Active Shape Analysis} Unlike the image domain, there are few methods using user interactions to aid 3D shape segmentation. One of the earliest techniques proposed in \cite{wang2012} asks the user to select pairs of segments between shapes to denote if they have the same or different segments. This technique is driven by an unsupervised method, so segmentation between shapes can be mismatched, and thus the user input to select the right shapes is crucial. 
Similarly, \cite{wu2014} asks the user to paint regions of the shapes for segment matching. Both methods tend to focus on smaller sets of shapes and use unsupervised methods to drive the segmentation. Recently, \cite{yi2016b} proposed a framework for annotating massive 3D shape dataset. They offer a crowd-sourcing application for annotators label a specified region which is used to train a conditional random field model. Once model predictions are obtained, the user would then be asked to verify the results by selecting all shapes that fail to have adequate annotations. While this technique shows good performance, fine details such as accurate segment boundaries can be difficult to achieve. Further, the user is only asked to verify results, and cannot fine tune 'almost acceptable' segmentations. These 'almost acceptable' segmentations then end up going through another round of model predictions or user labeling. Finally, this approach is a labeling pipeline, where a full pass provides a single segment for a dataset. Therefore, datasets with many distinct segments require many full passes to achieve a complete segmentation. Our proposed method alleviates all of these problems. By making use of a fast and robust deep learning model and effective refinement tools we provide high quality full segmentations in efficient times.

\textbf{Entropy uses in Geometry Processing} Entropy is a measure of uncertainty. It can be used to predict the probability of an event given some information. Entropy was first used in 3D geometry processing by \cite{page2003}, where it was used to estimate how much information was contained in a 3D surface. More recently, entropy has also been used for shape simplification \cite{xing2010}, shape compression \cite{lee2014} and to estimate the Saliency of a 3D shape \cite{limper2016}. As we have no way of analyzing our model evaluation, we explored using entropy to rank the segmentation predictions, which to our knowledge has not been explored before for 3D shape analysis.


\section{Framework Overview}
\label{sec:framework_overview}
Our active learning framework aims to produce a full segmentation for every shape in a given dataset, whilst minimizing the users' manual efforts. The input to our framework is a collection of manifold 3D shapes of any size $S$, from a specific category (e.g. aircraft), and a set of pre-defined segment identifiers $L$ (e.g. wings, engines, body, stabilizer). With our framework, users guide the selection of shapes for labeling, interact with the prediction of a deep learning model, and verify the segmentation quality of each shape. The framework will then produce an output of per-face labels for each shape, indicating which face belongs to which segment. The pipeline for our framework is shown in Figure~\ref{fig:pipeline}.

The pipeline consists of several components. Each component is inspired by the observations of existing problems, leading to our contribution of a fast and reliable active framework.

The framework is driven by a deep learning model, which predicts the labels for faces given feature descriptors (See section~\ref{sec:features} for feature details). As it is a supervised system, initial training data is required. This is obtained by the user manually segmenting several shapes. To aid the user in this task, the system will first suggest a small subset of shapes for the user to label, this is done by clustering global shape descriptors (See Section~\ref{sec:lfd}). This subset aims to well represent the dataset, to aid model generalization early on.

When manually labeling the dataset, the system offers the user many effective tools to speed up the process while still maintaining the high segmentation quality. These tools include robust over-segmentation and effective painting utilities (Section~\ref{sec:manual-seg}) and automatic boundary refinement (Section~\ref{sec:boundary-smoother}).

Once the user confirms the labels for any shape the deep learning model considers them as ground truth for training. So with the initial subset fully labeled, a model will be trained and used to predict results for the remaining shapes in the set (See Section~\ref{sec:deep-learning}). These results are displayed in our interactive table, which can be ordered in many useful ways (See Section~\ref{sec:table-ordering}). These different ordering methods allow the user to quickly see which shapes are correct and add them to the completed set (to be used for future model training). Alternatively, the table also quickly shows which shapes the model struggled with, these can then be manually labeled to make the model more generalized and increase overall quality.

The deep learning model is designed to be both quick and give high quality results, as such, the above steps can be repeated in quick succession to achieve a strong and generalized model. This can be used to quickly and effectively segment the entire dataset, with the user requiring less and less input per iteration (See Figure~\ref{fig:timing_comparison}).

\begin{figure}[t]
	\centering
	\includegraphics[width=\columnwidth]{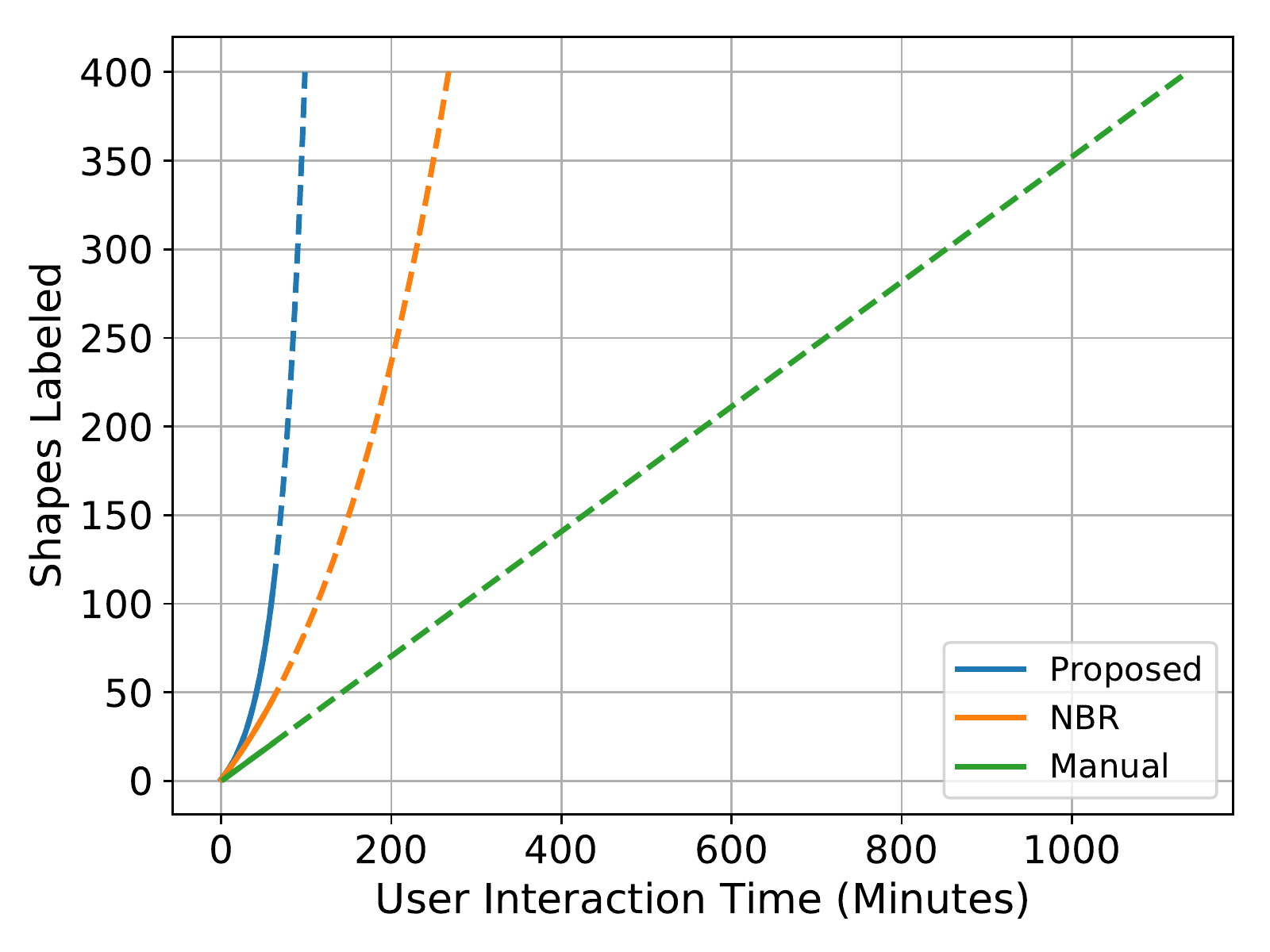}
	\caption{This graph shows the number of segmented shapes increasing as users use the system. The proposed line represents our system with all features enabled, the NBS (No Boundary Smoother) line represents our system with the boundary refinement feature disabled and the manual represents a system where only painting is enabled. The data making up the solid lines are from our 1 hour user studies, while the dashed lines are interpolated from the user study data. The figure shows that our system considerable speeds up shape segmentation when compared to a manual painting approach}
	\label{fig:timing_comparison}
\end{figure}

\section{Methodology}
\label{sec:methodology}

This section details all of the functions and tools provided by our active learning framework, in order to minimize user input and time, while still keeping a high segmentation quality. 

\subsection{Input Datasets}
\label{sec:input}
The input to our framework is a dataset of 3D shapes. As our method makes effective use of both geodesic distance and graph traversal, the input shapes must be manifold. This could be an issue for newer datasets, such as ShapeNet \cite{chang2015}, however, in this paper we also propose a robust re-meshing method for such datasets. This method has over 90\% success rate from our experiments, and has allowed us to provide a manifold sub-set of ShapeNet, with the ability to easily map back to the original shapes (See Section~\ref{sec:shapenet}). Each dataset $S$ consists of $n$ shapes, $S = \{s_1, s_2, ... s_{n-1} s_n\}$, where the $i$th shape $s_i = \{F, V, E\}$ is made up of faces $F$, vertices $V$ and edges $E$.

\subsection{Feature Extraction}
\label{sec:features}
As a pre-processing step we compute several features which help drive the framework. Specifically, we compute 3 face-level features and 1 shape-level feature. We use \acf{SC} \cite{belongie2002} and \acf{SI} \cite{johnson1999} as input for our deep learning architecture, which acts as a dual-branch ConvNet. This independently compresses both features down then combining them for classification (See Section~\ref{sec:deep-learning}). These features are both represented as 16x16 2D histograms, where \ac{SC} contains both geodesic distance and uniform angle \cite{kalogerakis2010}, and \ac{SI} contains information of shape vertex locations around a face. We also utilize the \acf{SDF} \cite{shapira2008}, which is used to aid our automatic boundary refinement process (See Section~\ref{sec:boundary-smoother}). 

Finally we compute \acp{LFD} \cite{chen2003} for each shape in the dataset. Similar to \cite{yi2017}, we extract multi-view snapshots of the shapes and then compute the \ac{HOG} features of those views. We concatenate the \ac{HOG} features of all views together and use that feature vector as the \ac{LFD}. We capture 20 views of the shape, and each \ac{HOG} feature is computed with 9 orientations, a cell size of [8, 8] and a block size of [2, 2]. This results in a 203760-dimension feature vector. We then embed the \ac{LFD} \ac{HOG} features from all shapes into a single space using \ac{PCA}, to make each feature vector 128-dimensions. These shape-level features are used as an embedded space for shape selection (See Section~\ref{sec:lfd})

\begin{figure}[t]
	\centering
	\includegraphics[width=\columnwidth]{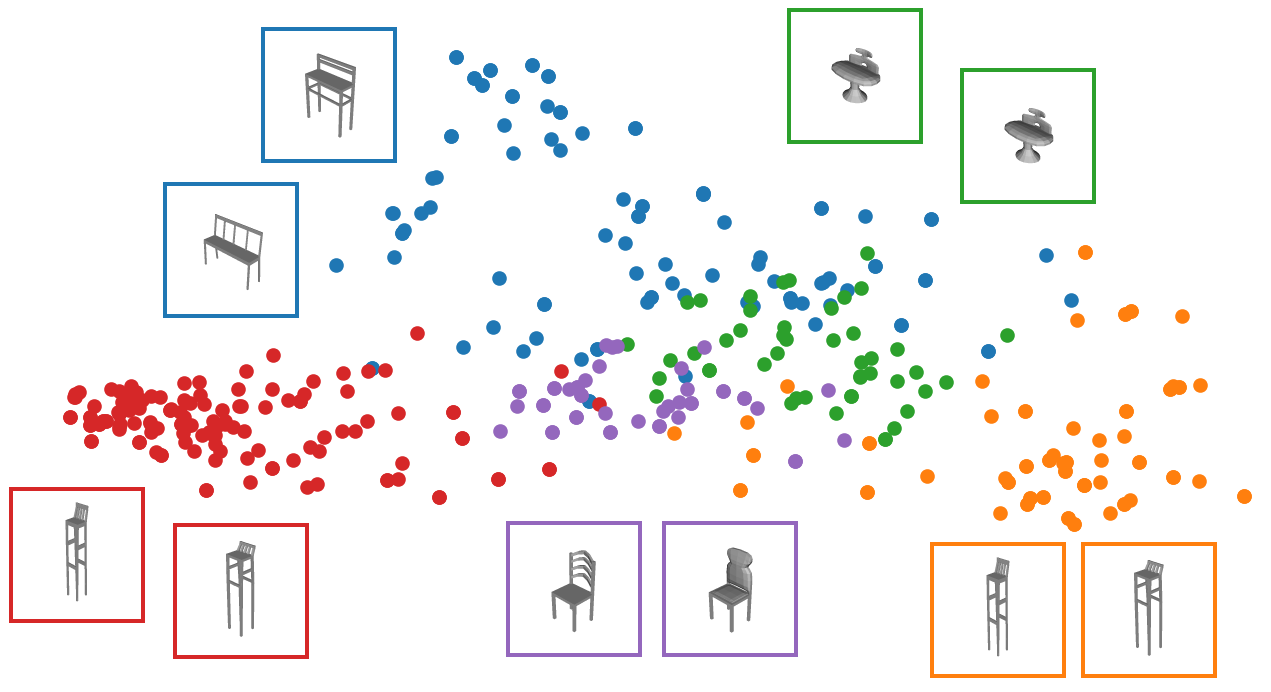}
	\caption{Shape embedding space from COSEG \cite{sidi2011} ChairsLarge dataset using 128-dimension \ac{LFD} \ac{HOG} features. Shapes displayed are the two closest shapes to the corresponding cluster centers.}
	\label{fig:lfd-embedding}
\end{figure}

\subsection{Initial Shape Selection}
\label{sec:lfd}
As our framework is driven by a deep learning architecture, in order to train it we first need some labeled data samples. One solution to this would be to let the user arbitrarily select some initial shapes from the dataset and manually label them. While this solution works to provide labeled training data, there is a chance that this data will not be well distributed throughout the dataset.

We address this by providing an embedded view of the data, which can then be clustered as desired. For a given number of clusters $k$, we compute k-means on the embedded \ac{LFD} \ac{HOG} features for the full dataset. Then, given the cluster labels $C^l$ and cluster centers $C^c$ we compute the $n$ closest shapes to each cluster center. These shapes are displayed to the user so they can select ones they wish to manually segment (Figure~\ref{fig:lfd-embedding}).

\subsection{Manual Segmentation Tools}
\label{sec:manual-seg}
Letting the use segment several shapes in a naive way is one possible way of obtaining an initial training set for the model. Such as letting them paint the entire shape from scratch or select segment boundaries with precise clicks. While this works for existing work \cite{yi2016b}, their work focuses primarily on single part labeling with little regard for segment boundaries. As we are focusing on full shape segmentation with additional care to preserve good segment boundaries, such a way of segmenting shapes manually would be too time consuming or produce poor results.

As such, our manual segmentation pipeline contains several useful tools to aid the user in quickly and effectively segmenting each shape. Here we outline the manual segmentation pipeline, showing useful tools that help in each step.

{\textbf{Shape Over-segmentation}} The first step in the pipeline is to assign segment labels to an over-segmentation of the shape. We provide two options for the over-segmentation, in most cases the shape can be segmented to an almost completed level using random walks segmentation \cite{lai2009}. For the other cases, we also offer a k-means clustering to give uniform patches across the shape. The outcome of either over-segmentation algorithm is a set of patches across the shape. These patches can quickly be assigned a segment label by the user so that they can move onto the refinement. The user does not have to give all patches a label, as the boundary smoothing algorithm (See Section~\ref{sec:boundary-smoother}) used to transition from this stage to the refinement stage will assign a label to any unlabeled patches.

{\textbf{Segmentation Refinement}}
At this stage, the shape is fully segmented, however, some modifications may be needed to achieve a good quality segmentation or to make segment boundaries acceptable. In this stage the user is able to 'paint' the shape to change the segment labels assigned to specific faces. There are several useful tools available in this stage:

\begin{itemize}
	\item {\textbf{Variable Sized Painting}} When painting a shape, a breadth first search algorithm is used to traverse it and assign the new label to faces it traverses. This algorithm is constrained by a adjustable radius (which is clearly shown to the user during painting). This allows for both large label corrections, or altering very fine details on boundaries.
	\item {\textbf{Paint Restrictions}} As a radius based breadth first search is used to paint the shape, it is possible that bounding sphere that the radius creates will pass over to parts of the shape the user does not wish to paint. To alleviate this, an angle based restriction can be enabled, which will compare the face normal of every traversed face to the normal of the face that was clicked. If the angle is greater than a user changeable threshold then  the face will not be painted. As many segment boundaries lie on concave parts of shapes, this can be very useful for quickly refining boundaries.
	\item {\textbf{Segment-wide Paint}} If an entire part of the shape is mislabeled (Sometimes the result of the deep learning algorithm), it can be quickly re-labeled to another segment by using this feature. All connected faces to the clicked face will be re-assigned the new label.
	\item {\textbf{Multiple Shape Views}} Quickly analyzing the quality of a segmentation is essential to minimize the time and effort needed by the user. For this, we allow for multiple views of the shape to be shown at once. This feature is best suited for quickly analyzing the output of the deep learning model, however, it can also be useful in the early stages of the pipeline.
\end{itemize}

A final feature, which is the most useful and powerful tool in this stage is the automatic boundary smoother, which is covered below in Section~\ref{sec:boundary-smoother}. By making use of all of these tools the user interaction time and effort can be cut down substantially while keeping the segmentation quality high, which is the key focus of this work.

\begin{figure}[t]
	\centering
	\null\hfill
	\begin{subfigure}[t]{0.24\columnwidth}
		\includegraphics[width=\columnwidth]{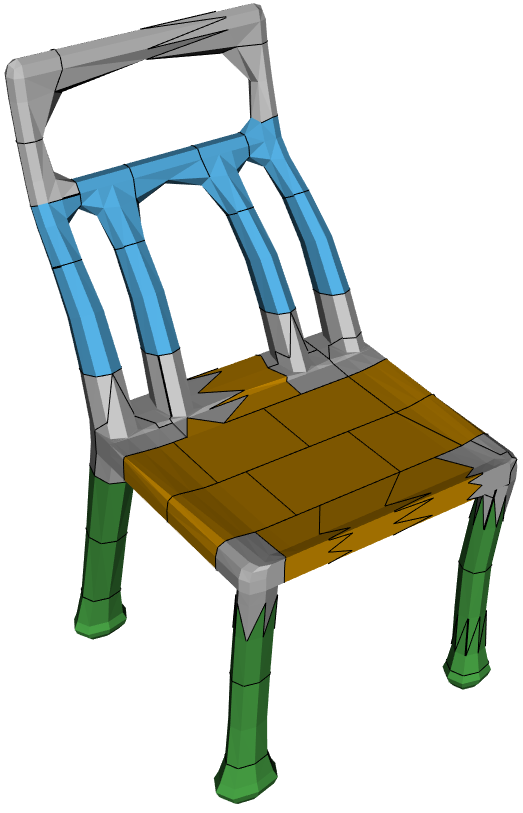}
		\caption{Patches}
	\end{subfigure}
	\hfill
	\begin{subfigure}[t]{0.24\columnwidth}
		\includegraphics[width=\columnwidth]{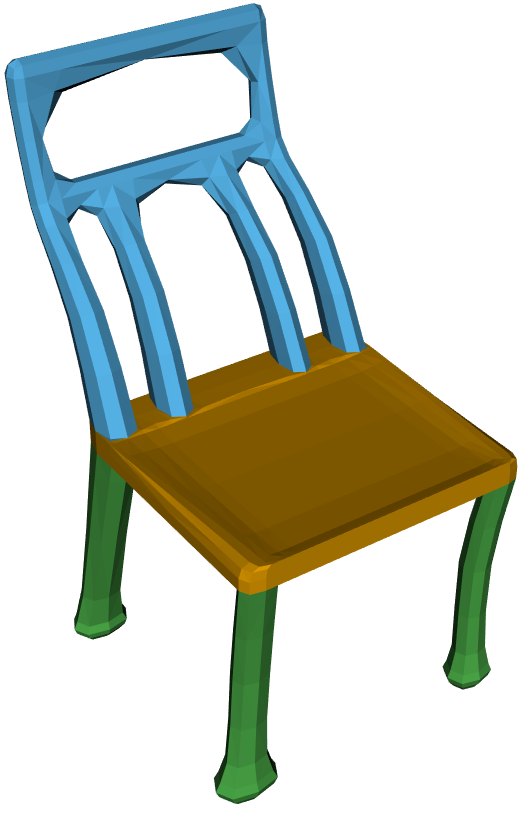}
		\caption{Result}
	\end{subfigure}
	\null\hfill
	\begin{subfigure}[t]{0.24\columnwidth}
		\includegraphics[width=\columnwidth]{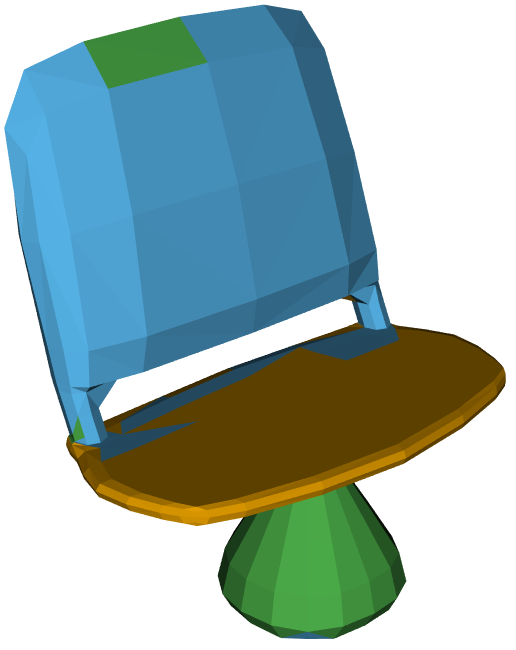}
		\caption{Predictions}
	\end{subfigure}
	\hfill
	\begin{subfigure}[t]{0.24\columnwidth}
		\includegraphics[width=\columnwidth]{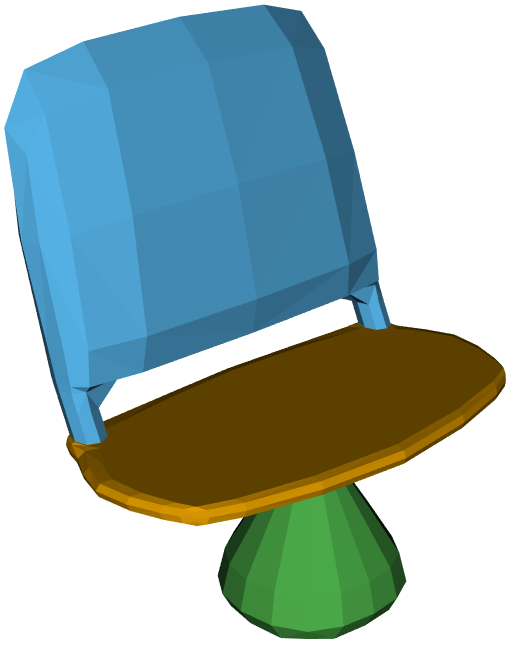}
		\caption{Result}
	\end{subfigure}
	\hfill\null
	\caption{Resulting segmentation from running our boundary refinement algorithm. The algorithm can take incomplete (gray input patches are considered unlabeled) patch segmentations (a) and return very good refined results (b). The algorithm also performs well when given model predictions (c) as the results show in (d)}
	\label{fig:bs-results}
\end{figure}

\begin{figure}[t]
	\centering
	\begin{subfigure}[t]{0.32\columnwidth}
		\includegraphics[width=\columnwidth]{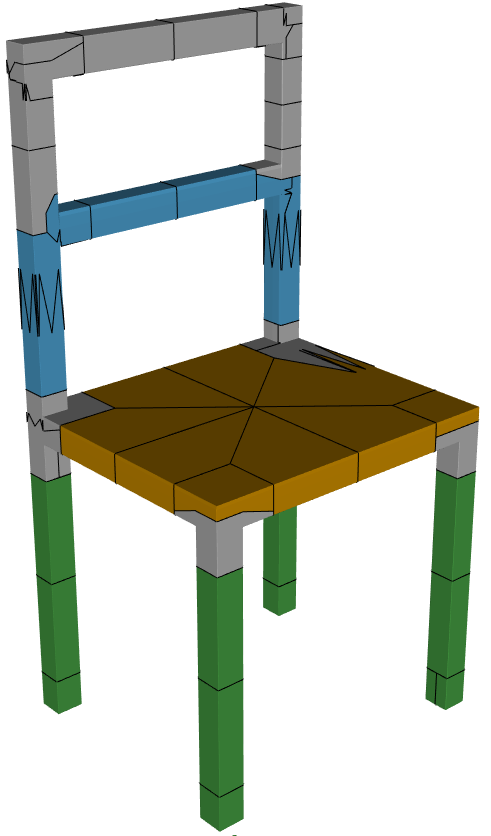}
		\caption{Input}
	\end{subfigure}
	\begin{subfigure}[t]{0.32\columnwidth}
		\includegraphics[width=\columnwidth]{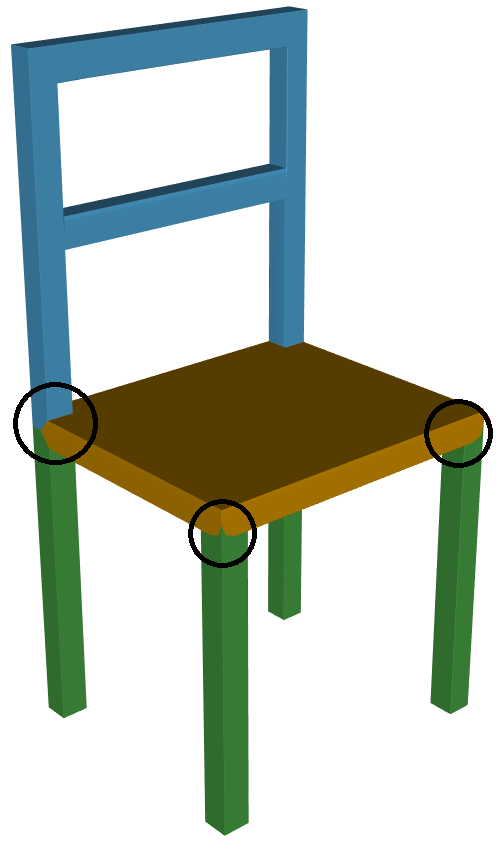}
		\caption{Result - without \ac{SDF}}
	\end{subfigure}
	\begin{subfigure}[t]{0.32\columnwidth}
		\includegraphics[width=\columnwidth]{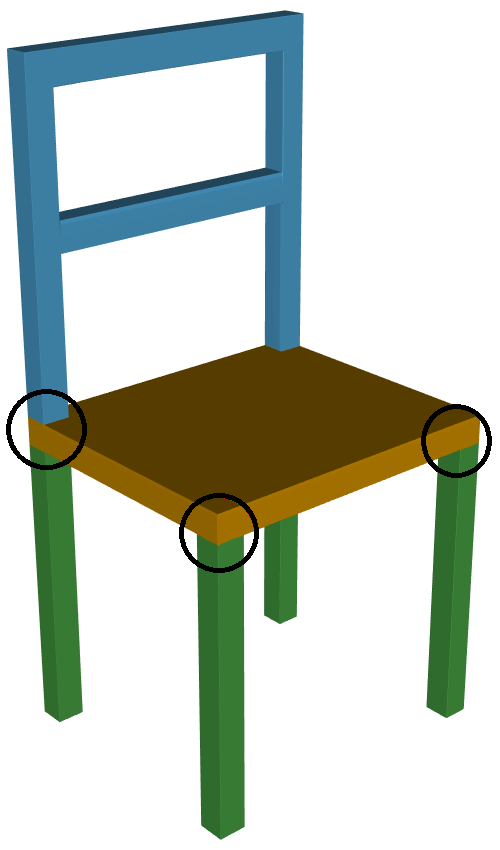}
		\caption{Result - with \ac{SDF}}
	\end{subfigure}
	\caption{Comparison of our boundary refinement algorithm with and without \ac{SDF} in the smoothness term. The segment boundaries (black circles) are poor when \ac{SDF} is not used (b), and very good when \ac{SDF} is used (c) }
	\label{fig:bs-sdf}
\end{figure}

\subsection{Automatic Boundary Refinement}
\label{sec:boundary-smoother}

To achieve a high level segmentation with smooth boundaries in an active system, typically, the user would have to spend time in a refinement stage. In this stage, they would typically fine tune the small details of boundaries to achieve the desired result. This task is both tedious and time consuming, so to alleviate this we introduce an automatic boundary refining algorithm.

As the shapes can be treated as a graph and we are interested in consistent boundaries, we make use of the multi-label alpha expansion algorithm \cite{boykov2001}. Given $F$, the set of all faces, and face $f \in F$. Let $N_f$ be the set of neighboring faces of $f$, we can then optimize the labels for all faces by solving:

\begin{equation}
\min_{l_f, f \in F} \sum_{f \in F} \xi_D (f, l_f) + \sum_{f \in F, f' \in N_f} \xi_S (f, f'),
\label{eqn:bs-gc}
\end{equation}
where $l_f$ is the label assigned to face $f$. The data term $\xi_D$, is computed as a weighted geodesic distance term estimates the probability of assigning label $l$ to face $f$. Specifically, we define a set of boundary edges $E^{l}_{b}$ for each $l \in L$, which is made of pairs of neighboring faces $\{u, v\} \in E$, where $u, v \in F$, and where $l_u \ne l_v$, and either $l_u = l$ or $l_v = l$. Given $E^l_b$, we compute the shortest distance, $d^l_f = Gdist(f, E^l_b)$ for all $f \in F$ and all $l \in L$, where $Gdist(\cdot, \cdot)$ is the geodesic distance. We then compute the data term $\xi_D$ as:

\begin{equation}
\xi_D (f, l) = \begin{cases}
1, & \text{if $d^l_f \geq \sigma$ and $l = l_f$}\\
0.5, & \text{if $d^l_f < \sigma$}\\
0, & \text{otherwise},
\end{cases} 
\end{equation}
where $\sigma$ is a threshold based on the bounding box of the shape (we empirically set this to 0.01 times the bounding diagonal). The smoothness term $\xi_S$, penalizes large curvature between adjacent faces and is given by:

\begin{equation}
\xi_S (f, f') = \pi - \theta_{ff'}
\label{eqn:bs-smooth}
\end{equation}
where $\theta_{ff'}$ is the dihedral angle between the normals of faces $f$ and $f'$. By solving Equation~\ref{eqn:bs-gc}, we obtain a new set of labels which are smoothed and aim to preserve boundaries. This functionality is useful in both transitioning from patches to painting, and also as a refinement method for the output of the deep learning model (see Figure~\ref{fig:bs-results}).

One drawback of our algorithm, is the inability to detect segment boundaries that do not lie on high curvature regions. The smoothness term $\xi_S$ is driven by face curvature, so will not detect when a boundary should lie on a low curvature surface. One solution to this comes from the observation that segment boundaries also typically lie on regions with a large change in thickness \cite{shapira2008}. With this observation in mind, we include implementation for a second boundary refinement algorithm which makes use of \ac{SDF}. This implementation modifies Equation~\ref{eqn:bs-smooth} to the following:

\begin{equation}
\xi_S (f, f') = \omega \cdot (\pi - \theta_{ff'}) + (1 - \omega) \cdot \delta_{ff'}
\label{eqn:bs-smooth-with-sdf}
\end{equation}
where $\delta_{ff'}$ is the absolute difference between the \ac{SDF} feature of faces $f$ and $f'$, and $\omega$ is a weight (between 0 and 1) which the user can change (default is 0.2). A comparison of both algorithms is shows in Figure~\ref{fig:bs-sdf}, which shows an example shape which would fail with the standard boundary refinement algorithm, but gives good results with the modified version which uses \ac{SDF}. The user can choose to use either of these algorithms as both are strong in different ways.

\subsection{Deep Learning Label Predictions}
\label{sec:deep-learning}

The core of our active learning pipeline is the deep learning model. By providing a small subset of the data, our fast and effective model will predict segmentations for the remainder of the dataset, removing the need for manual labeling from scratch.

\begin{figure}[t]
	\centering
	\includegraphics[width=\columnwidth]{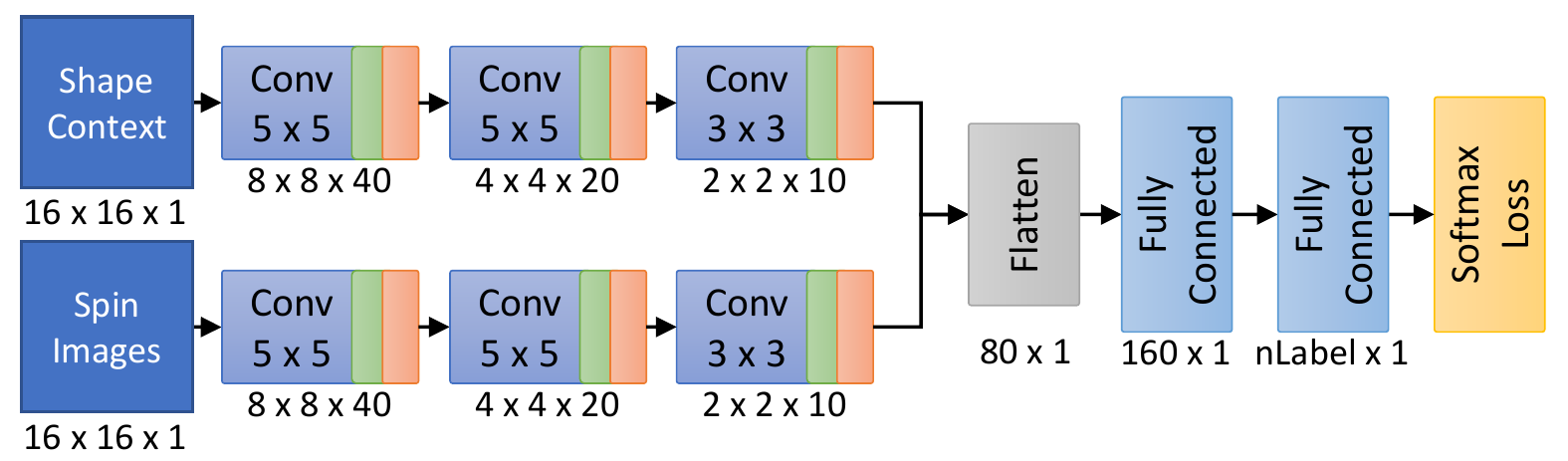}
	\caption{Architecture of our deep learning model. A Conv block consists of a convolution layer with leaky \ac{ReLU} \cite{xu15icmlworkshop} activation ($\alpha$ = 0.2), then a 2x2 max pooling layer with a stride of 2. The numbers underneath each layer represent the output size. The architecture separately compresses both input features before combining them to compute predicted class labels.}
	\label{fig:architecture}
\end{figure}

We have designed our deep learning architecture with both speed and performance in mind. The model must be quick to train and evaluate so that the user is not waiting for long periods, but the model must also be accurate to further minimize the users input and time spent. With this in mind we designed a novel convolutional neural network to separately compress two features (\ac{SC} and \ac{SI}) and non-linearly combine them for label predictions (See Figure~\ref{fig:architecture} for the architecture)

Previous work has shown that geometric features can be very useful in making a deep learning model both accurate and generalized \cite{guo2015, george2018}, so we opted to use two of the strongest features in our architecture and separately compress them using convolutional and pooling layers. We did this because both features are 2D histograms, so convolving over them in a 2D space is a logical way of compressing the size of the features while maintaining as much of information as possible. Once compressed, the two features are flattened down to a feature vector and concatenated. We then pass this feature vector through a small fully connected network to obtain the final predictions (Figure~\ref{fig:architecture}).

\begin{figure}[t]
	\centering
	\includegraphics[width=\columnwidth]{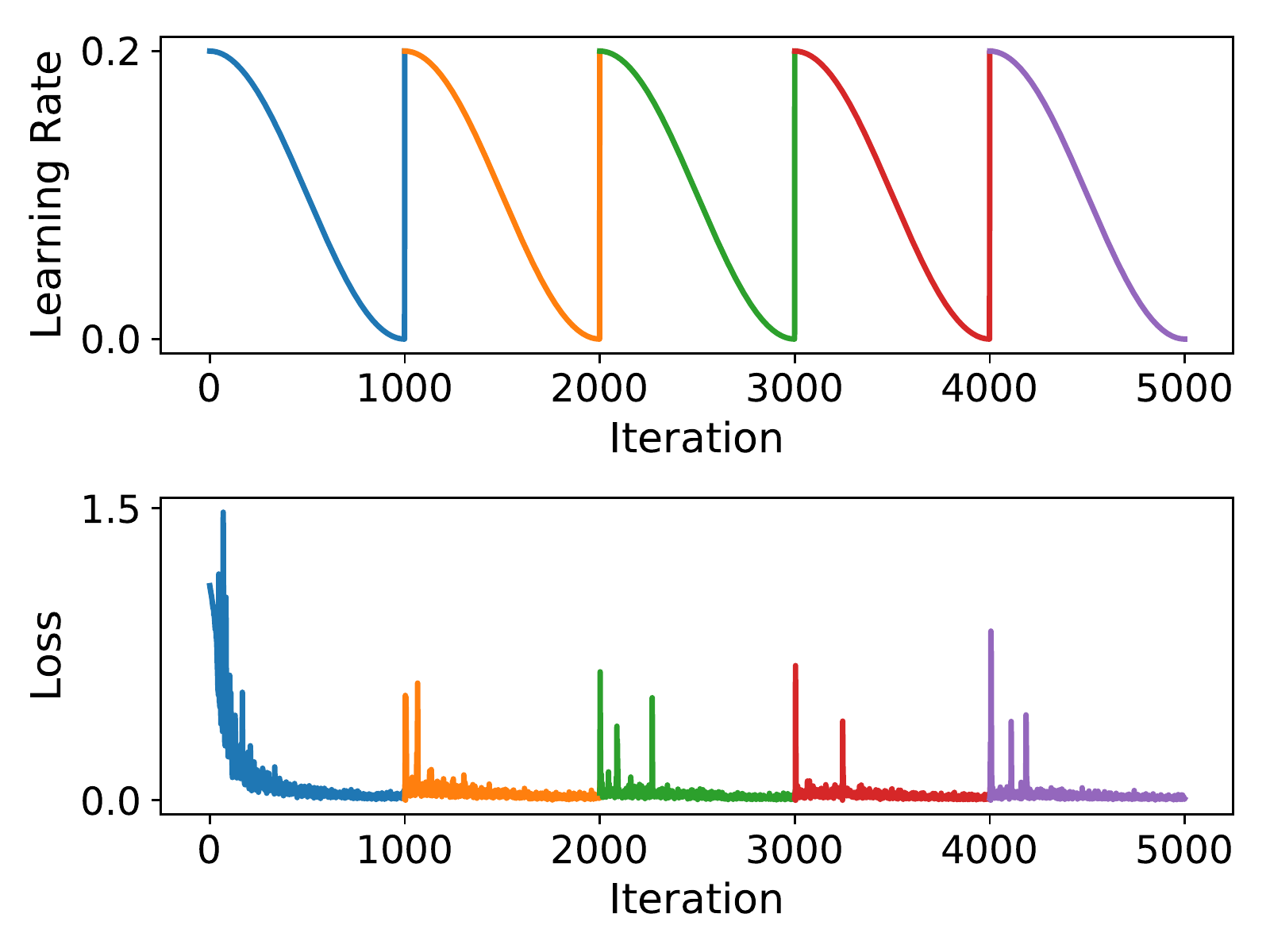}
	\caption{Learning rate and loss plotted as the model is trained. Different snapshots are shown in different colors. Each time the learning rate resets the model the optimizer is forced out of a local minima making the loss spike}
	\label{fig:lr_loss}
\end{figure}

We chose to train our models using a snapshot ensemble \cite{huang2017} learning scheme. This was because it allows for multiple models to be trained in the same amount of time, increasing the ability for the model to generalize. Empirically, we chose to train our networks using the RMSProp \cite{tieleman2012} optimizer. In our experiments we train each model for 5000 iterations, $T$, and save 5 snapshots, $M$, of the model weights. We employ the same learning rate function as proposed by \cite{loshchilov2016}: 

\begin{equation}
\alpha(t) = \frac{\alpha_0}{2} \left(\cos\left(\frac{\pi\mod(t - 1, \lceil T / M \rceil)}{\lceil T / M \rceil}\right) + 1\right)
\label{eqn:lr}
\end{equation}
where $\alpha_0$ is the initial learning rate (We set $\alpha_0 = 0.01)$. This gives a learning rate $\alpha$ for any given $t < T$. The learning rate resets $M$ times so that the model can escape local minima giving a more generalized model \cite{huang2017} (See Figure~\ref{fig:lr_loss}). As more shapes are labeled and the number of training samples grows, we opted to uniformly sample batches (Each with 512 samples) from the entire pool of data. This allows us to fix the number of iterations and still train generalized models. This stops the model from taking an increasing amount of time to train each time new shapes are labeled.

Once an ensemble of trained models has been obtained the remaining shapes in the dataset can be evaluated. The features for all the faces of a shape is passed through each network in the ensemble, we then extract the label probabilities and average them across all ensembles, giving $\bm{p}$. The label with the highest probability is considered the segment label for a given face, we call this the predicted segmentation. 

{\textbf{Graph Cut Refinement}} In addition, we also compute a refined segmentation by again making use of multi-label alpha expansion \cite{boykov2001} by solving:

\begin{equation}
\min_{l_f, f \in F} \sum_{f \in F} \phi_D (f, l_f) + \lambda \sum_{f \in F, f' \in N_f} \phi_S (f, f'),
\label{eqn:gc}
\end{equation}
where $\lambda$ is a non-negative constant used to balance the terms and $\phi_D (f, l_f) = -\log(\bm{p}_f(l_f))$ penalizes low probability of assigning a label $l_f$. The second term, $\phi_S = -\log(\pi - \theta_{ff'})$, penalizes adjacent faces which form concavities, where $\theta_{ff'}$ is the dihedral angle between the face normals of faces $f$ and $f'$. This refinement technique has been used in recent work (\cite{guo2015, sidi2011,george2018}), but is slightly modified in comparison. The output of the refinement is a segment label per face, we call this the refined segmentation.

\begin{figure}[t]
	\centering
	\begin{subfigure}[t]{\columnwidth}
		\includegraphics[width=\textwidth]{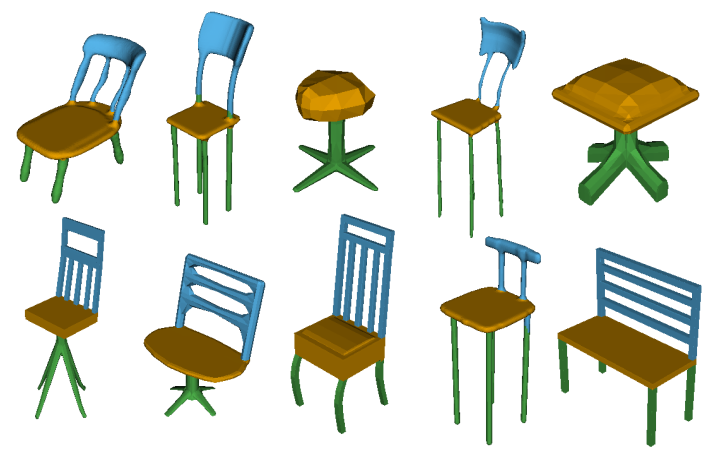}
		\caption{High ranking shapes}
	\end{subfigure}
	
	\begin{subfigure}[t]{\columnwidth}
		\includegraphics[width=\textwidth]{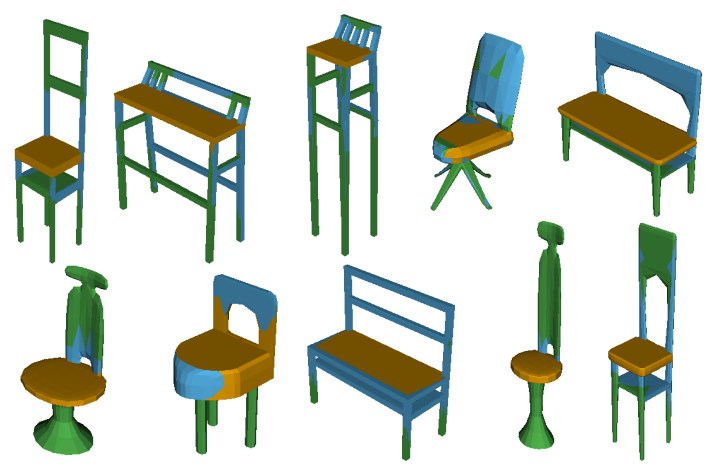}
		\caption{Low ranking shapes}
	\end{subfigure}
	\caption{Visual comparison of high (a) and low (b) ranking shapes when ranking according to entropy.}
	\label{fig:entropy}
\end{figure}

\subsection{Effective Table Ordering}
\label{sec:table-ordering}

Each time evaluation of the model is completed, the user is presented with the results. At this stage, they have several options of how to view the them. These options are spread across three menus, where one option per menu is selected at a given time.

\textbf{Displayed Segmentation} The first menu governs which segmentation is displayed on each model in the table. There are two options to choose from; predicted segmentation and refined segmentation. These are the outputs from the deep learning model and the graph cut refinement, respectively.

\textbf{Ordering Algorithm} The second menu controls how the table is ordered. There are two options; no order and entropy. Given the probability matrix $\bm{p}$, of shape $S$, 
the entropy score is computed as:
\begin{equation}
E_S = \sum_{\bm{p}_f \in \bm{p}} \frac{\sum_{\bm{p}^l_f \in \bm{p}_f} -\bm{p}^l_f \log(\bm{p}^l_f)}{n^S_f}
\label{eqn:entropy}
\end{equation}
where $n^S_f$ is the number of faces in shape $S$, $\bm{p}_f$ are all probabilities for face $f$, and $\bm{p}^l_f$ is the probability of face $f$ being assigned label $l$. As we don't have ground truth labels to evaluate the performance of the remaining unlabeled shapes, we needed another measure for ranking the shapes. Entropy is a measure of uncertainty of a probability distribution, therefore it is a natural alternative. For a shape we measure the entropy of each face and then average it across all other faces. This provides a score, which we then use to order the whole dataset.

%

\begin{figure}[t]
	\centering
	\begin{subfigure}[t]{.61\columnwidth}
		\includegraphics[width=\textwidth]{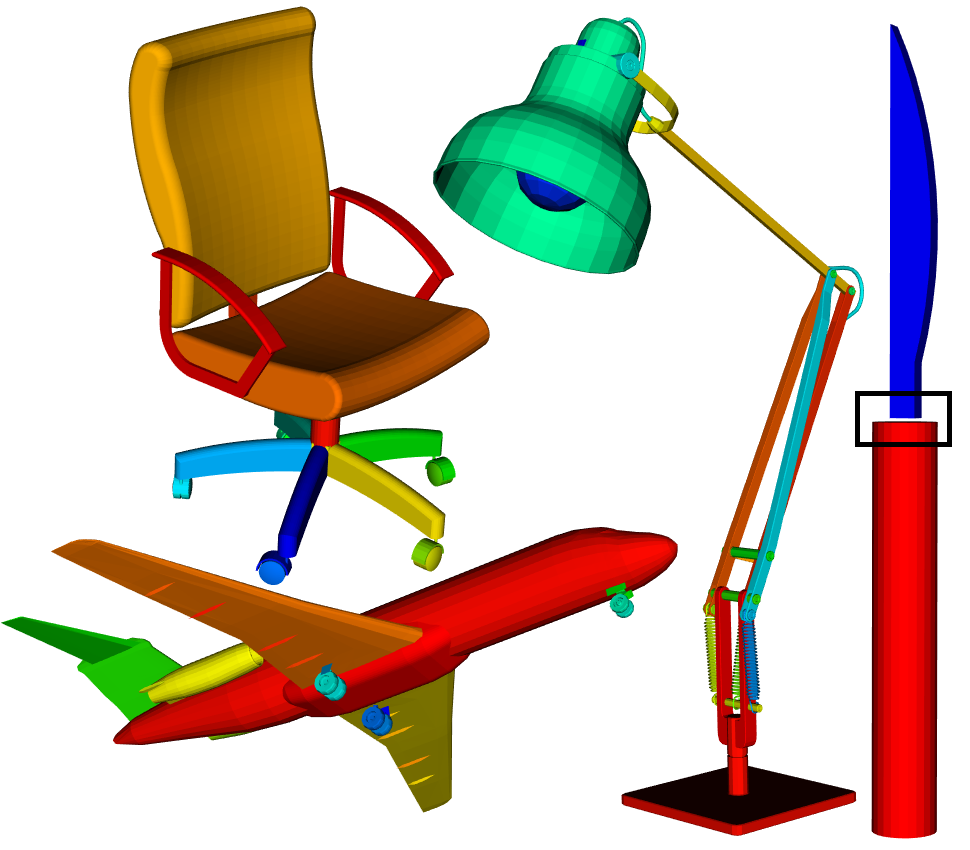}
		\caption{Multiple components}
	\end{subfigure}
	\begin{subfigure}[t]{.35\columnwidth}
		\includegraphics[width=\textwidth]{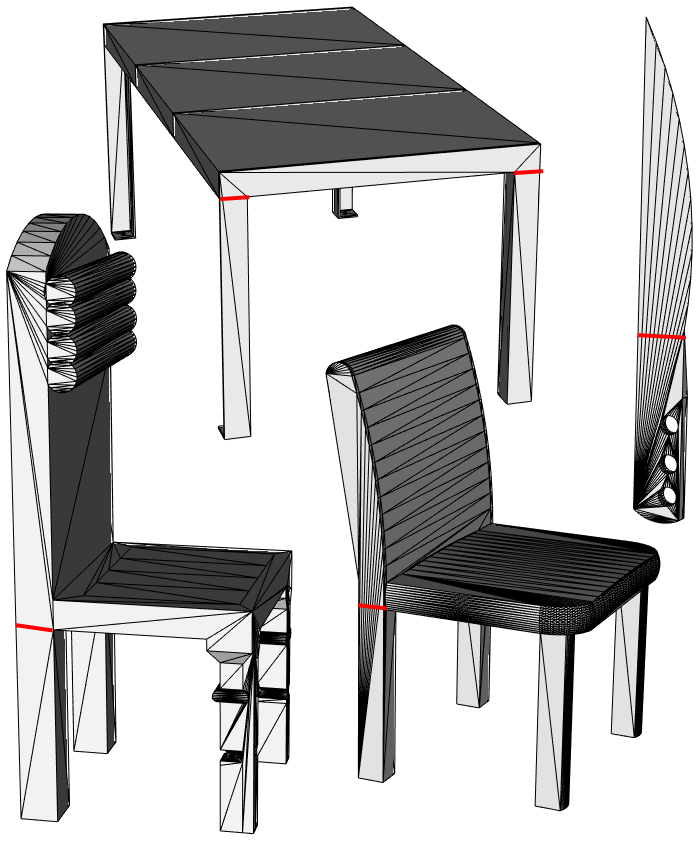}
		\caption{Low resolution}
	\end{subfigure}
	\caption{Examples from ShapeNet where shapes have multiple components (a), and are low resolution (b). Different components are denoted by different colors and red lines show where segment boundaries would lie. We also show a case where two components have a significant gap between them (black box, (a)).}
	\label{fig:shape-net-issues}
\end{figure}

\textbf{Order Direction} The final option controls if the data is presented in ascending (worst to best) or descending (best to worst) order.

Effectively using these different ordering methods can greatly reduce the time needed to label a dataset. The ordering is based off the models predictions, and manually refining shapes which the model had trouble segmenting makes the model generalize quicker to the rest of the dataset. A visual comparison entropy ranking is shown in Figure~\ref{fig:entropy}, which allows the user to quickly see shapes the model is good and bad at segmenting.

\subsection{ShapeNet Re-meshing}
\label{sec:shapenet}

ShapeNet \cite{chang2015} is a massive online repository of 3d shapes used frequently in shape retrieval and matching techniques. The repository contains thousands of 3D shapes from dozens of shape categories giving shape analysis algorithms the potential to be evaluated on widely diverse datasets. Recently, shape segmentation techniques have also begun using ShapeNet datasets for benchmarking their proposed algorithms \cite{qi2016, evangelos2017}, this was initiated by the ground truth labels from \cite{yi2016b}.

However, due to the nature of such a large repository of shapes, many of the shapes are not manifold and consist of many disconnected regions or even polygon soups. Due to this, many techniques have shifted to a point cloud driven algorithm, which sacrifices much of the information that can be obtained from a manifold shape. Furthermore, any mesh driven technique is limited to point cloud ground truth segmentations (provided by \cite{yi2016b}), as extracting a meaningful segmentation on the provided shapes is challenging due to low face counts, poor geometry and miscellaneous parts (See Figure~\ref{fig:shape-net-issues}).

While certain issues can be rectified by simple re-meshing (face sub-division, vertex merging etc.), this would only make a small percentage of the shapes in ShapeNet datasets manifold with a reasonable face count. The remaining shapes require more sophisticated techniques to become manifold. Due to this we combine existing re-meshing techniques into a pipeline, which can successfully make the majority of ShapeNet manifold and with reasonable face counts. The re-meshing pipeline is as follows for a given input shape:

\begin{figure}[t]
	\centering
	\includegraphics[width=\columnwidth]{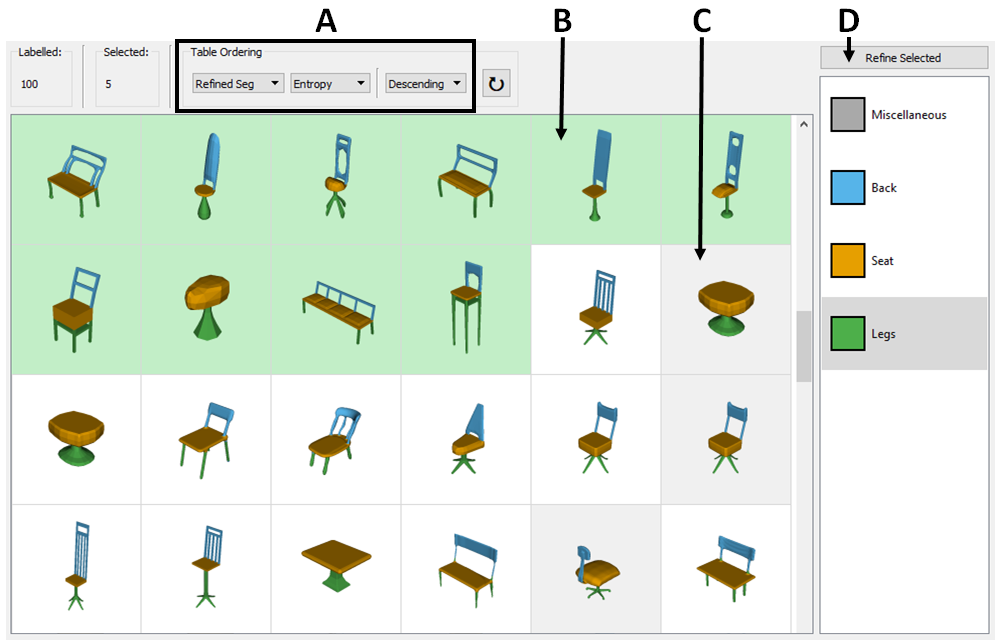}
	\caption{Table interface of our system. Allows for quick analysis of the entire dataset with an effective ordering method. \textbf{A} Table ordering (Section~\ref{sec:table-ordering}). \textbf{B} Shapes shown in green have full segmentation and have been user verified. These are used to train the deep learning model. \textbf{C} The table allows for selection of shapes for manual refinement and verification. \textbf{D} Visualize the selected shapes in the annotation interface (Figure~\ref{fig:interface_paint}).}
	\label{fig:interface_table}
\end{figure}

\begin{enumerate}
	\item For a 3D grid of fixed size, extract the distances from geometry points to grid points, essentially voxelising the shape.
	\item Pass voxelised shape through a contour filter to generate an isosurface.
	\item Extract the largest component from the isosurface, removing any internal cavities that may have been created.
	\item Assert that the new shape surface is manifold and geometrically similar to the original shape by comparing \ac{LFD} \ac{HOG} features.
	\item Decimate the new shape to several different face counts (50k, 20k 10k 5k) asserting manifoldness throughout. This allows us to provide high, medium and low quality re-meshing.
\end{enumerate}

\begin{figure}[!t]
	\centering
	\includegraphics[width=\columnwidth]{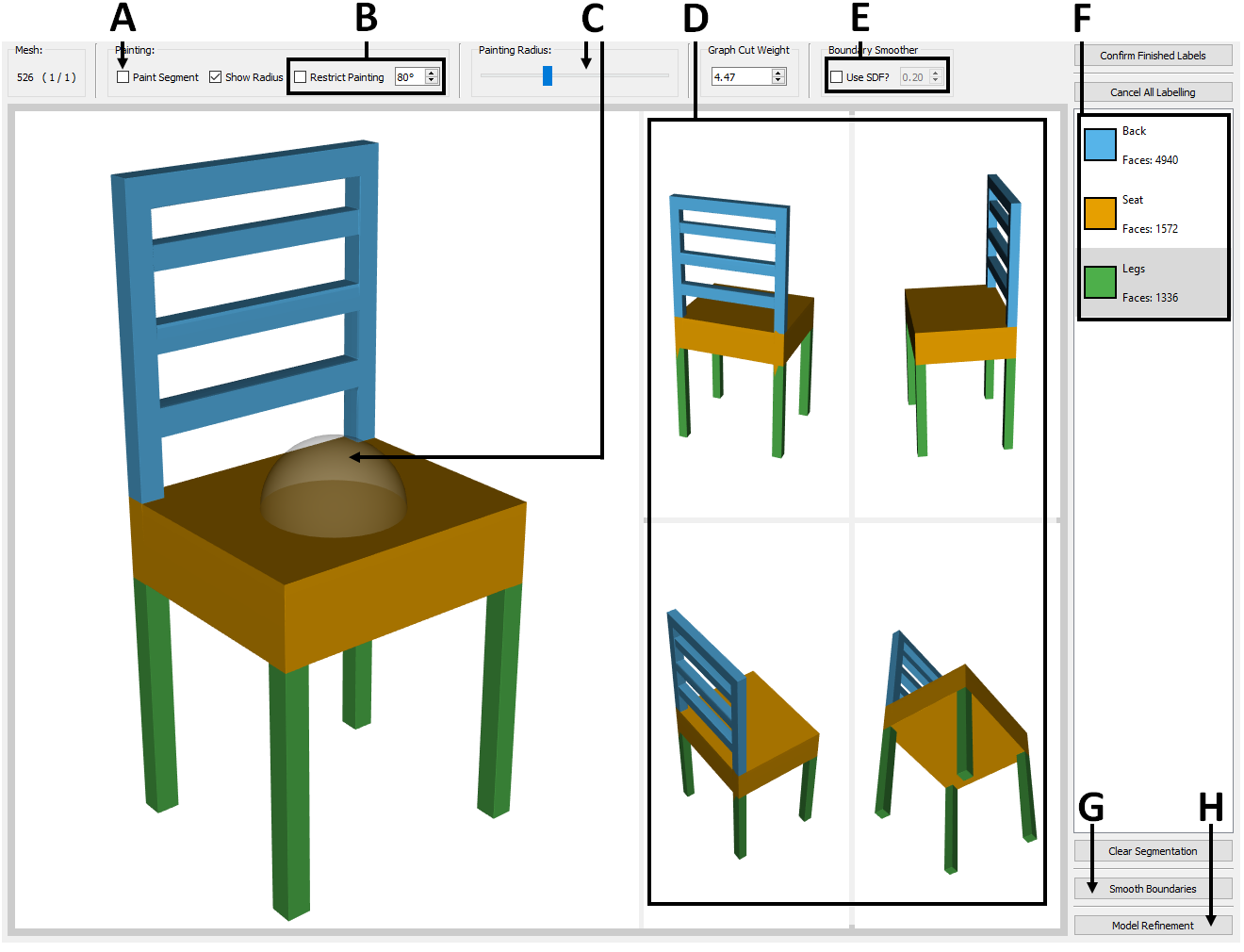}
	\label{fig:interface_paint}
	\caption{Annotation Interface of our system, where the user can label or refine a subset of shapes. \textbf{A} Segment wide paint (Section~\ref{sec:manual-seg}). \textbf{B} Painting restriction (Section~\ref{sec:manual-seg}). \textbf{C} Paint radius with visual indicator. \textbf{D} Multiple shape views for quick segmentation analysis (Section~\ref{sec:manual-seg}). \textbf{E} Weight of the SDF influence on the boundary refinement. \textbf{F} Segment names, colors and face counts. Selected (gray) segment will be assigned to faces when painting. \textbf{G} Boundary refinement (Section~\ref{sec:boundary-smoother}). \textbf{H} Model refinement (Section~\ref{sec:deep-learning})}
	\label{fig:interface_paint}
\end{figure}

The output of this pipeline is a set of manifold shapes with varying resolutions. These shapes can then be used on with any existing shape analysis pipeline, and are still compatible with the available ground truth segmentations via nearest neighbor matching. Any shapes that fails the asserts throughout the pipeline are passed through again with different tunable parameters (grid resolution, contour value), or removed from the dataset if all parameter permutations are exhausted.


\begin{table*}[t]
	\newcommand{\tbf}{\textbf}
	\newcommand{\mc}{\multicolumn{3}{c}}
	{\resizebox{\textwidth}{!}{    
			\begin{tabular}{@{}L{4cm} @{}C{1.4cm} @{}C{1.4cm} @{}C{1.4cm} @{}C{1.4cm} @{}C{1.4cm} @{}C{1.4cm} @{}C{1.4cm} @{}C{1.4cm} @{}C{1.4cm}@{}} 		
				\toprule
				& \mc{\tbf{No Refinement}} 							& \mc{\tbf{Graph Cut Refinement}}					& \mc{\tbf{Boundary Refinement}} 					\\
				& \tbf{Chairs} 	& \tbf{Vases}	& \tbf{Aliens}		& \tbf{Chairs} 	& \tbf{Vases}	& \tbf{Aliens}		& \tbf{Chairs} 	& \tbf{Vases}	& \tbf{Aliens}		\\ \toprule 
				& 92.19			& 89.98     	& 90.87     		& 96.76			& 92.73     	& 94.62     		& 96.66			& 92.80     	& 94.56     		\\ \midrule 
				\tbf{10 Snapshots (5)} 			& 91.73			& 89.96     	& 90.64     		& 96.67			& 92.34     	& 94.70     		& 96.79			& 92.78     	& 94.66     		\\ \midrule 
				\tbf{5 Snapshots} 				& \tbf{92.68}	& \tbf{90.19}  	& 90.91     		& \tbf{97.18}	& \tbf{92.75}  	& \tbf{95.02}  		& \tbf{97.03}	& \tbf{92.81}  	& \tbf{94.78}  		\\ \midrule 
				\tbf{3 Snapshots} 				& 92.27			& 90.01     	& \tbf{90.99}  		& 97.10			& 92.53     	& 94.82     		& 96.93			& 92.74     	& 94.73     		\\ \midrule 
				\tbf{1 Snapshot} 				& 91.43			& 89.14     	& 90.14     		& 96.63			& 91.57     	& 94.28     		& 96.93			& 91.91     	& 94.67     		\\ \midrule 
				\tbf{Fixed Learning Rate} 		& 79.49			& 87.37     	& 86.22     		& 83.41			& 90.19     	& 90.79     		& 83.99			& 90.73     	& 91.50     		\\ \midrule 
				\tbf{Decaying Learning Rate} 	& 87.34			& 86.42     	& 88.05     		& 93.93			& 89.42     	& 92.92     		& 94.45			& 89.83     	& 93.63     		\\ \bottomrule 
	\end{tabular}}}
	\captionof{table}{5-fold cross validation on the COSEG large datasets \cite{sidi2011} using different learning schemes and refinement techniques. (5) denotes only the last 5 snapshots were used for evaluation \cite{huang2017}. \tbf{Bold} values denote the highest accuracy for the set and refinement method}
	\label{table:ensemble_results}
\end{table*}

\section{Interface and Program Flow}
\label{sec:interface}

We provide a system with many useful tools for interactively segmenting a dataset of shapes. When using the system, the user is presented with two main interfaces, shown in Figures~\ref{fig:interface_table} and \ref{fig:interface_paint}. These interfaces dynamically change depending on which stage in the programs pipeline (Figure~\ref{fig:pipeline}), the user is currently at. Here we outline the program flow from each interface to the tools provided as the user progresses through the pipeline.

{\textbf{New Dataset}} This is the entry point of the system, where the user will be presented with the table interface showing all shapes in the dataset. At this point, the user will be able to initialize the different segments the dataset will contain. From here, the user needs to select starting shapes to manually segment. They have two options for this; arbitrarily pick them from the table, or use our initial shape selection tool (Section~\ref{sec:lfd}).

{\textbf{Coarse Segmentation}} Given the subset of selected shapes, the user is now tasked with manual segmentation, this is done in the annotation interface. This is the part of the pipeline which requires the most user time, as such, we provide many options. To quickly assign coarse labeling each shape is over-segmented, we provide two options for this (Section~\ref{sec:manual-seg} Shape Over-segmentation), allowing the user control over how many patches are generated. The user then assigns segment labels to the patches by clicking a segment with a specified label. They do not need to label all patches, and can transition to the next stage by using the boundary refinement algorithm to label the remaining segments and smooth the boundaries.

{\textbf{Segmentation Refinement}} In this stage the annotation interface changes to allow for 'painting' (shown in Figure~\ref{fig:interface_paint}). This is where the segmentation of a shape is completed to the users satisfaction. The boundary refinement algorithm can be used as often as needed to automatically adjust boundaries, and the user can then fix any small segmentation defects that exist. The user can mark the shape as complete (shown by a green boarder around the shape) and move on to another shape. Once all shapes in the subset are completed they are then stored as ground truth, and marked in green on the table interface.

{\textbf{Model Training and Evaluation}} Once any number of shapes are segmented, the deep learning model can be trained and evaluated. This can be done at any time and allows for the table to be ordered much more effectively.

{\textbf{Selecting the next subset}} Just like starting with a new dataset, the user can arbitrarily pick shapes from the table, or use our initial shape selection tool (Section~\ref{sec:lfd}). In addition, the table can now be ordered to rank the shapes according entropy (Section~\ref{sec:table-ordering}). The table can also be used to display the predicted or refined segmentation (Section~\ref{sec:deep-learning}). This can be useful, as the user can see shapes that the model has correctly segmented and quickly confirm them. Also, by using the ranking, the user can see shapes that the model couldn't segment well, the can then select these as part of the next subset. From this stage the user can either segment the subset from nothing by using {\textbf{Coarse Segmentation}}, or refine the predicted or refined segmentation using {\textbf{Segmentation Refinement}}.

The above program flow iterates and as the user confirms more shape segmentations, the model has access to more training data and better generalizes, reducing the future interaction effort (Figure~\ref{fig:timing_comparison}).


\section{Results and Discussions}
\label{sec:results}

In this section we will evaluate our interactive system. There are several key components that make up our system, here we will provide experiments and results carried out to evaluate the individual components.

\subsection{Deep Learning Model}

While any appropriate deep learning model or classification algorithm could be used with our system, this work also showcases a novel deep learning architecture, which is both fast and effective for 3D shape segmentation. To evaluate our model and design choices we include results from several experiments. These evaluate the performance of the deep learning model, the performance of the refinement techniques and support the use of ensemble based learning. 

Firstly, we evaluate the choice of an ensemble based learning scheme. We performed 5-fold cross validation of the 3 large COSEG datasets \cite{sidi2011} with varying numbers of snapshots. For comparison, we also performed the same experiments with a fixed learning rate and decaying learning rate. We include these as examples of typical learning rate values for model training. The starting learning rate in all experiments was 0.01. All ensemble experiments used Equation~\ref{eqn:lr} to update the learning rate. The decaying learning rate experiment reduced the learning rate by a factor of 10 at 50\% and again at 75\% of the training process. The results are shown in the No Refinement columns of Table~\ref{table:ensemble_results}. As the columns shows, using a snapshot learning scheme consistently improves results when compared to fixed and decaying learning rate. There is also a considerable increase in accuracy when only using a single snapshot, which shows that the cosine learning function (Equation~\ref{eqn:lr}) alone improves the quality of the trained model. Finally, the results show that in the majority of cases 5 snapshots give the best performance increase. We the remaining experiments use this.

\begin{table}[t]
	\newcommand{\tbf}{\textbf}
	{\resizebox{\columnwidth}{!}{                                                  
			\begin{tabular}{@{}L{1.8cm} @{}C{1.5cm} C{1.5cm} C{1.5cm} C{1.5cm}@{}} 
				\toprule
				& \tbf{PCA \& NN}		& \tbf{2D CNN} 			& \tbf{1D CNN} 		& \tbf{Proposed} 	\\ \toprule
				\tbf{Airplane} 	& 92.53       			& 94.56                	& 96.52				& 95.22				\\ \midrule
				\tbf{Ant}      	& 95.15       			& 97.55                	& 98.75 			& 98.75 			\\ \midrule
				\tbf{Armadillo}	& 87.79       			& 90.90                	& 93.74 			& 94.99 			\\ \midrule
				\tbf{Bird}     	& 88.20       			& 86.20                	& 91.67   			& 88.64   			\\ \midrule
				\tbf{Chair}    	& 95.61      			& 97.07       			& 98.41   			& 97.61   			\\ \midrule
				\tbf{Cup}      	& 97.82       			& 98.95				    & 99.73   			& 98.12   			\\ \midrule
				\tbf{Fish}    	& 95.31       			& 96.16          		& 96.44 			& 96.43 			\\ \midrule
				\tbf{Fourleg}  	& 82.32      			& 81.91                	& 86.74 			& 84.55 			\\ \midrule
				\tbf{Glasses}   & 96.42 				& 96.95					& 97.09				& 98.10				\\ \midrule
				\tbf{Hand}      & 70.49       			& 82.47                	& 89.81 			& 88.21 			\\ \midrule
				\tbf{Human}  	& 81.45       			& 88.90			        & 89.81				& 90.66				\\ \midrule
				\tbf{Octopus}   & 96.52       			& 98.50			 		& 98.63				& 98.71				\\ \midrule
				\tbf{Plier}     & 91.53       			& 94.54                	& 95.61 			& 95.32 			\\ \midrule
				\tbf{Table}     & 99.17 				& 99.29					& 99.55				& 98.99				\\ \midrule
				\tbf{Teddy}    	& 98.20 				& 98.18          		& 98.49				& 98.57				\\ \midrule
				\tbf{Vase}     	& 80.24       			& 82.81			      	& 85.75   			& 82.87   			\\ \midrule			                                                            
				\tbf{Average}   & 90.61       			& 92.79  				& 94.80 			& 94.11 			\\ \bottomrule
	\end{tabular}}}                                                                   
	\captionof{table}{Leave-one-out cross validation on the \ac{PSB} dataset \cite{chen2009}. \ac{PCA} \& \ac{NN}, 2D \ac{CNN} \cite{guo2015} and 1D \ac{CNN} results from \cite{george2018}}
	\label{table:psb_results}
\end{table}

Next we evaluate the accuracy of our deep learning architecture. We devised two sets of experiments; leave-one-out cross validation on the PSB dataset \cite{chen2009} and 5-fold cross validation on the COSEG dataset \cite{sidi2011}. We used the recent work from \cite{george2018} as a comparison as they provide results from several feature driven deep learning architecture, they also include results for the 2D CNN from \cite{guo2015}. Tables~\ref{table:psb_results} and \ref{table:coseg_results} show the results of the experiments. As our model architecture was designed with speed in mind we compare against existing models that can be trained quickly (\ac{PCA} \& \ac{NN}, 2D \ac{CNN}). Table~\ref{table:psb_results} shows that our proposed architecture has a significant increase (3.5\%) over \ac{PCA} \& \ac{NN} and even a moderate increase (1.3\%) over a more sophisticated 2D CNN \cite{guo2015}. Also when compared to a much bigger network \cite{george2018}, which uses more than 4x more input features and takes considerably longer to train, our proposed method's performance is still within 1\% (on average). Similarly, in Table~\ref{table:coseg_results} our proposed method again out performs the 2D CNN, with a much more considerable increase (4\%) when looking at large datasets.

\begin{table}[t]
	\newcommand{\tbf}{\textbf}
	{\resizebox{\columnwidth}{!}{    
			\begin{tabular}{@{}L{2.5cm} @{}C{1.75cm} @{}C{1.75cm} @{}C{1.75cm}@{}}
				\toprule
				& \tbf{2D CNN} 	& \tbf{1D CNN}	& \tbf{Proposed}		\\ \toprule    
				\tbf{Candelabra}			& 91.55 		& 93.58     	& 91.35     		\\ \midrule
				\tbf{Chairs}   				& 93.48			& 97.75     	& 94.82     		\\ \midrule
				\tbf{Fourleg}				& 90.75			& 94.12     	& 92.40     		\\ \midrule
				\tbf{Goblets}  				& 92.79			& 97.80     	& 87.40     		\\ \midrule
				\tbf{Guitars}  				& 97.04			& 98.03     	& 96.23     		\\ \midrule
				\tbf{Irons}    				& 80.90			& 89.89     	& 85.96     		\\ \midrule
				\tbf{Lamps}    				& 81.52			& 86.74     	& 91.44     		\\ \midrule
				\tbf{Vases}  				& 89.42			& 92.47     	& 86.08     		\\ \midrule
				\tbf{Average}				& 89.68			& 93.80			& 90.71				\\ \bottomrule \\ \toprule
				\tbf{VasesLarge} 			& 87.57			& 95.88     	& 92.81     		\\ \midrule 
				\tbf{ChairsLarge}  			& 92.68			& 97.71     	& 97.18     		\\ \midrule
				\tbf{AliensLarge}  			& 91.93			& 97.84     	& 95.02     		\\ \midrule  
				\tbf{Average}				& 90.73			& 97.14			& 95.00				\\ \bottomrule
	\end{tabular}}}
	\captionof{table}{5-fold cross validation on the COSEG dataset \cite{sidi2011}. 2D \ac{CNN} \cite{guo2015} and 1D \ac{CNN} results from \cite{george2018}}
	\label{table:coseg_results}
\end{table}

Finally, we evaluate our refinement techniques, Graph Cut Refinement (End of Section~\ref{sec:deep-learning}) and Automatic Boundary Refinement (Section~\ref{sec:boundary-smoother}). As the outputs of our ensemble experiments were reported without any refinement, we pass these outputs through our two refinement algorithms. The Graph Cut Refinement and Boundary Refinement columns of Table~\ref{table:ensemble_results} show the results of the experiments. As the columns show, both techniques show a considerable increase in accuracy compared to the un-refined results. The refined results also further support the use of 5 snapshot based ensembles, providing the highest accuracy across all results. One further observation is that the boundary refinement algorithm can be executed iteratively as the data term is based on the position of segment boundaries, which will change between runs. The results shown are after a single run of the boundary refinement, so it is possible for further improvements in the results using an iterative approach.

\begin{figure*}[t]
	\newcommand{\size}{0.32}
	{\centering
		\begin{tabular}{@{}c@{}c@{}c@{}}
			\includegraphics[width=\size\textwidth]{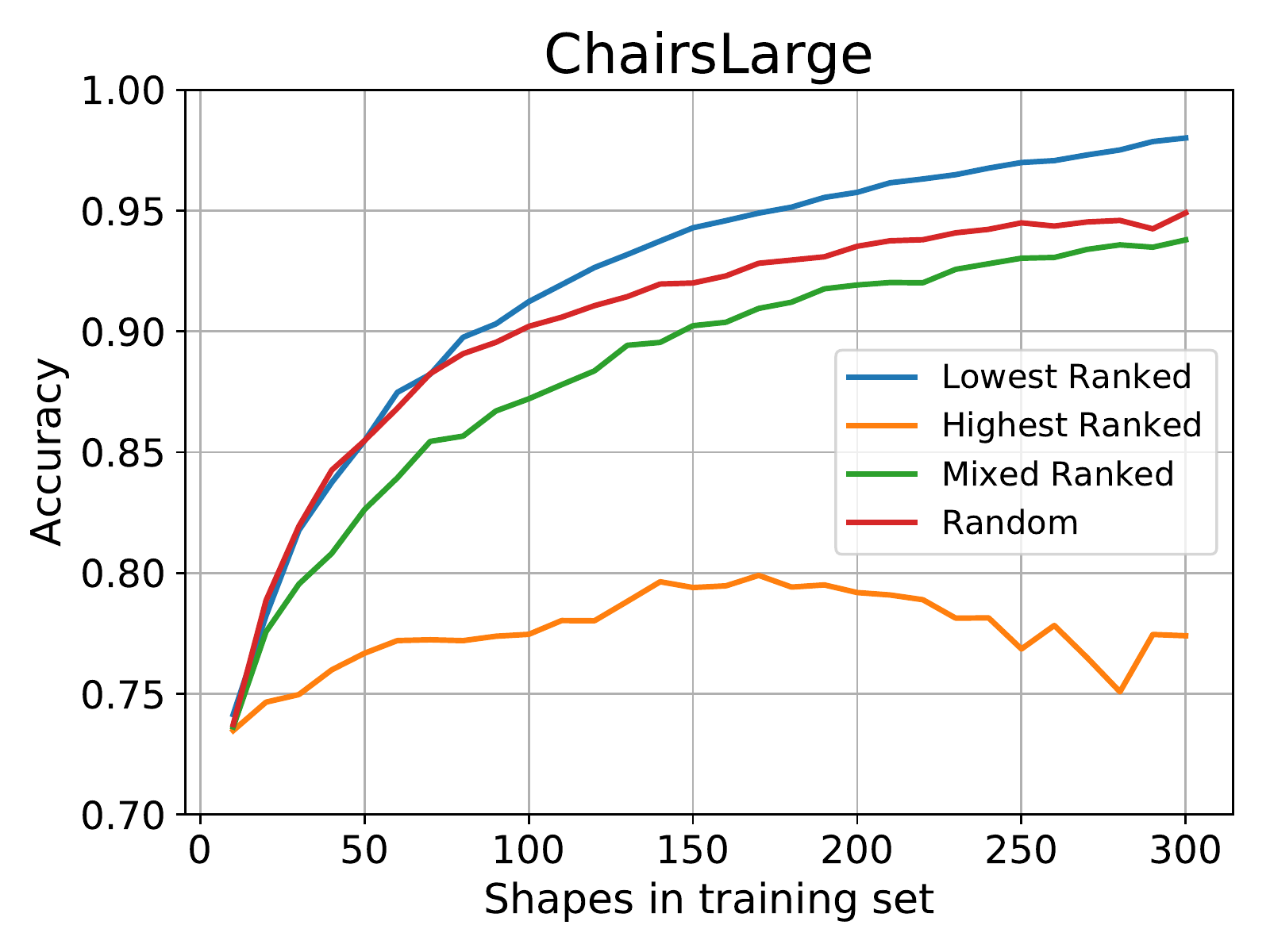} 	&
			\includegraphics[width=\size\textwidth]{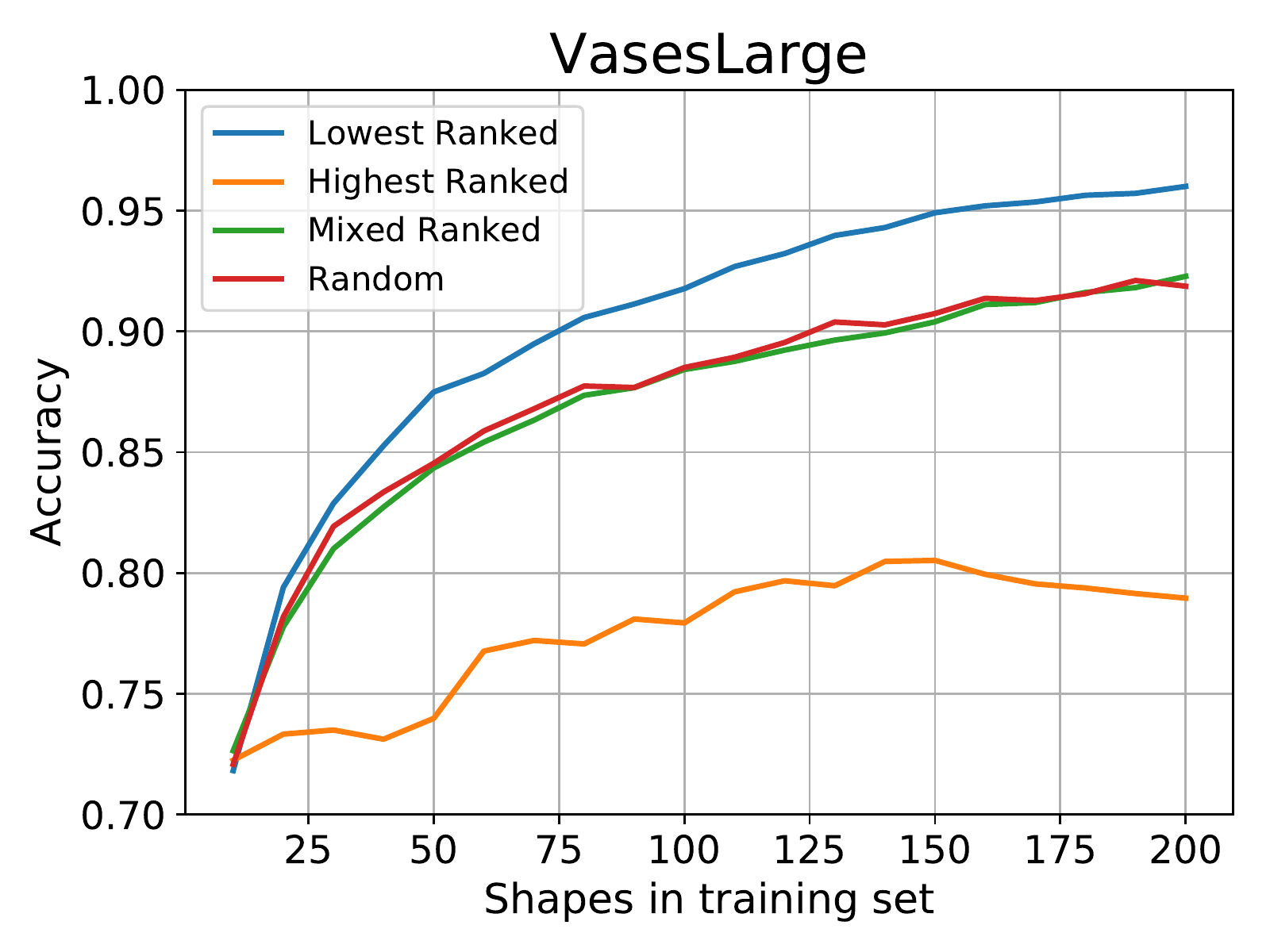}		&
			\includegraphics[width=\size\textwidth]{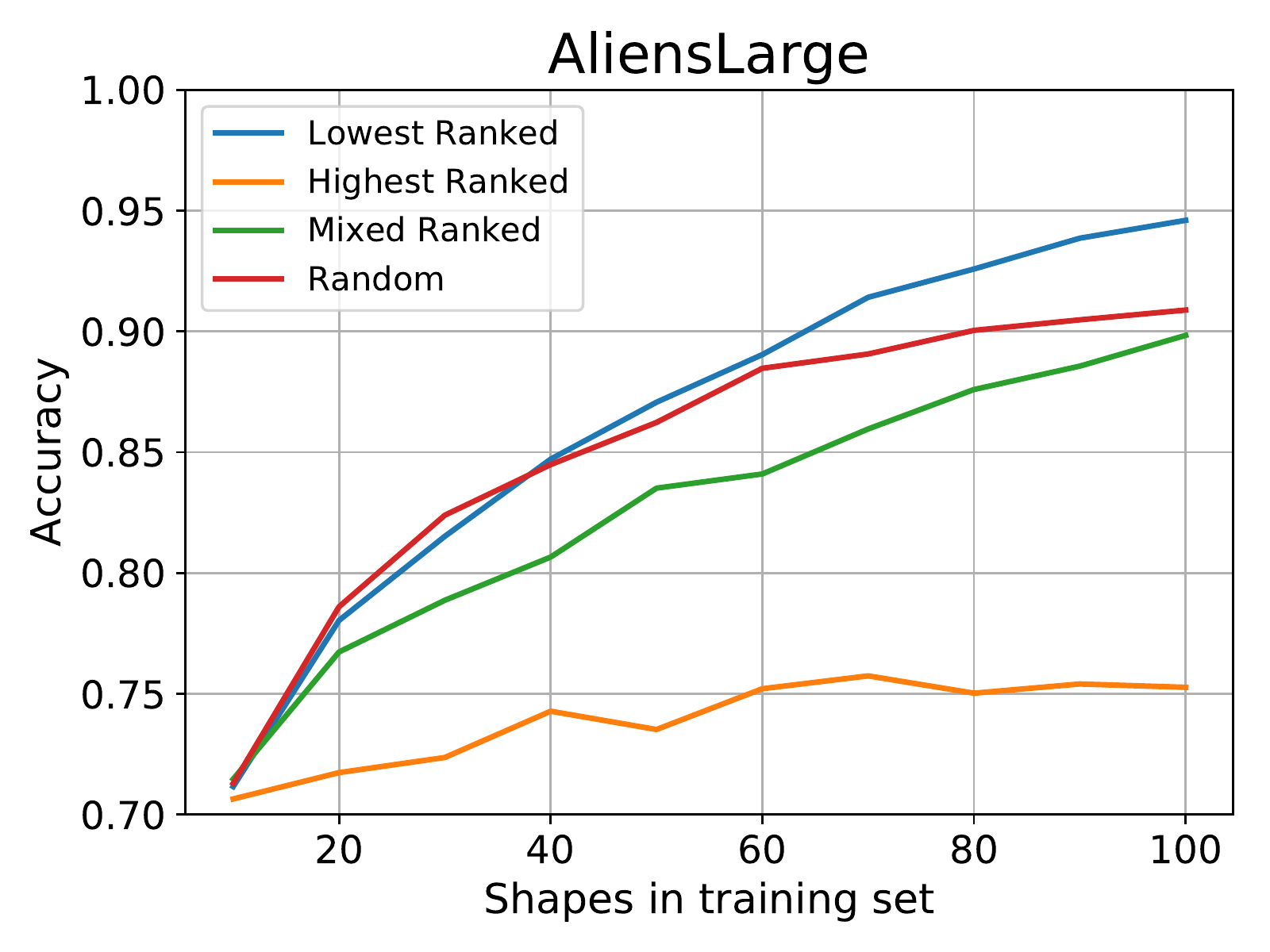}	\\
			\multicolumn{3}{c}{\begin{subfigure}{0.8\textwidth}\caption{Validation Set}\end{subfigure}} 	\\
			\includegraphics[width=\size\textwidth]{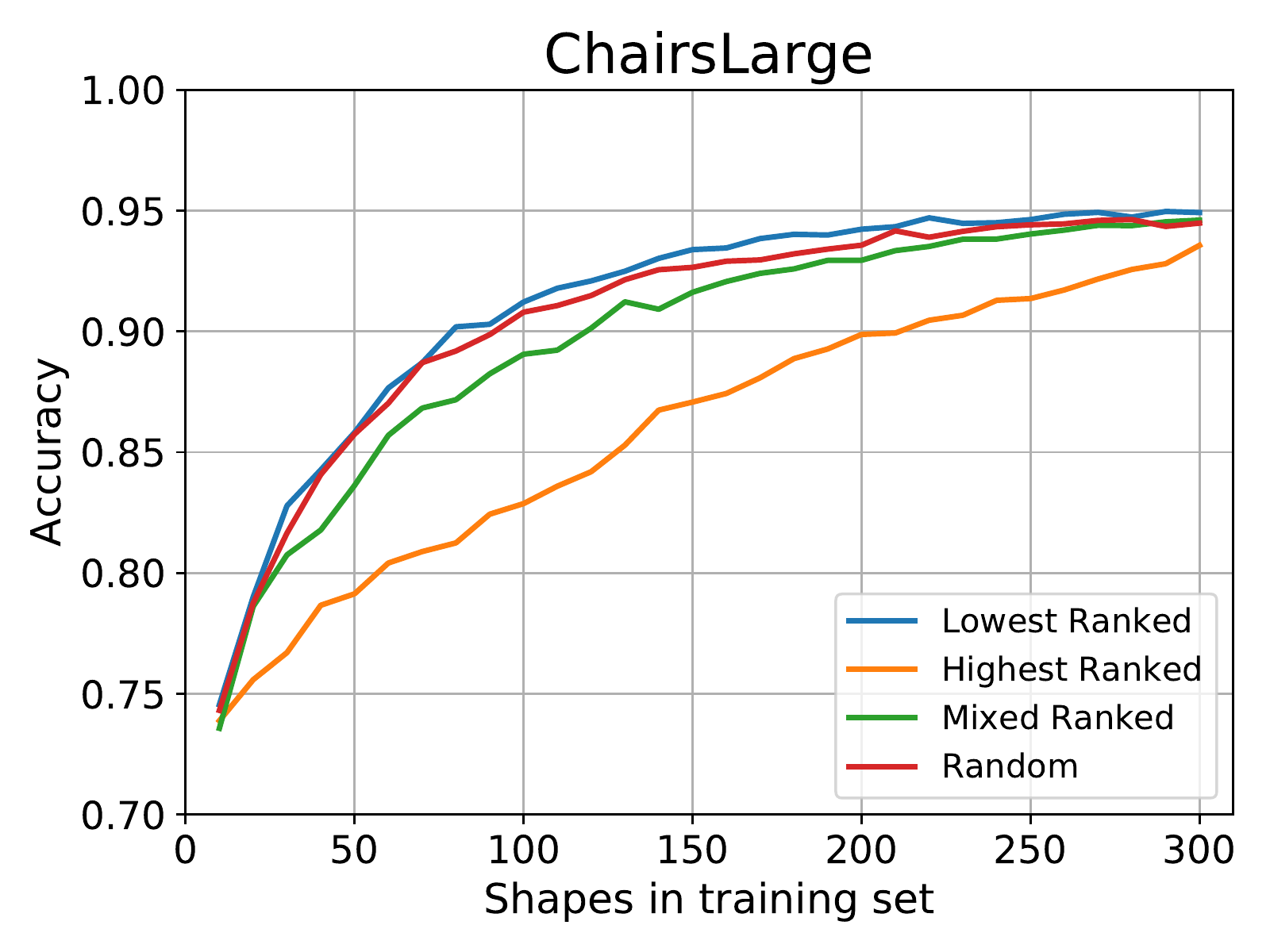}		&
			\includegraphics[width=\size\textwidth]{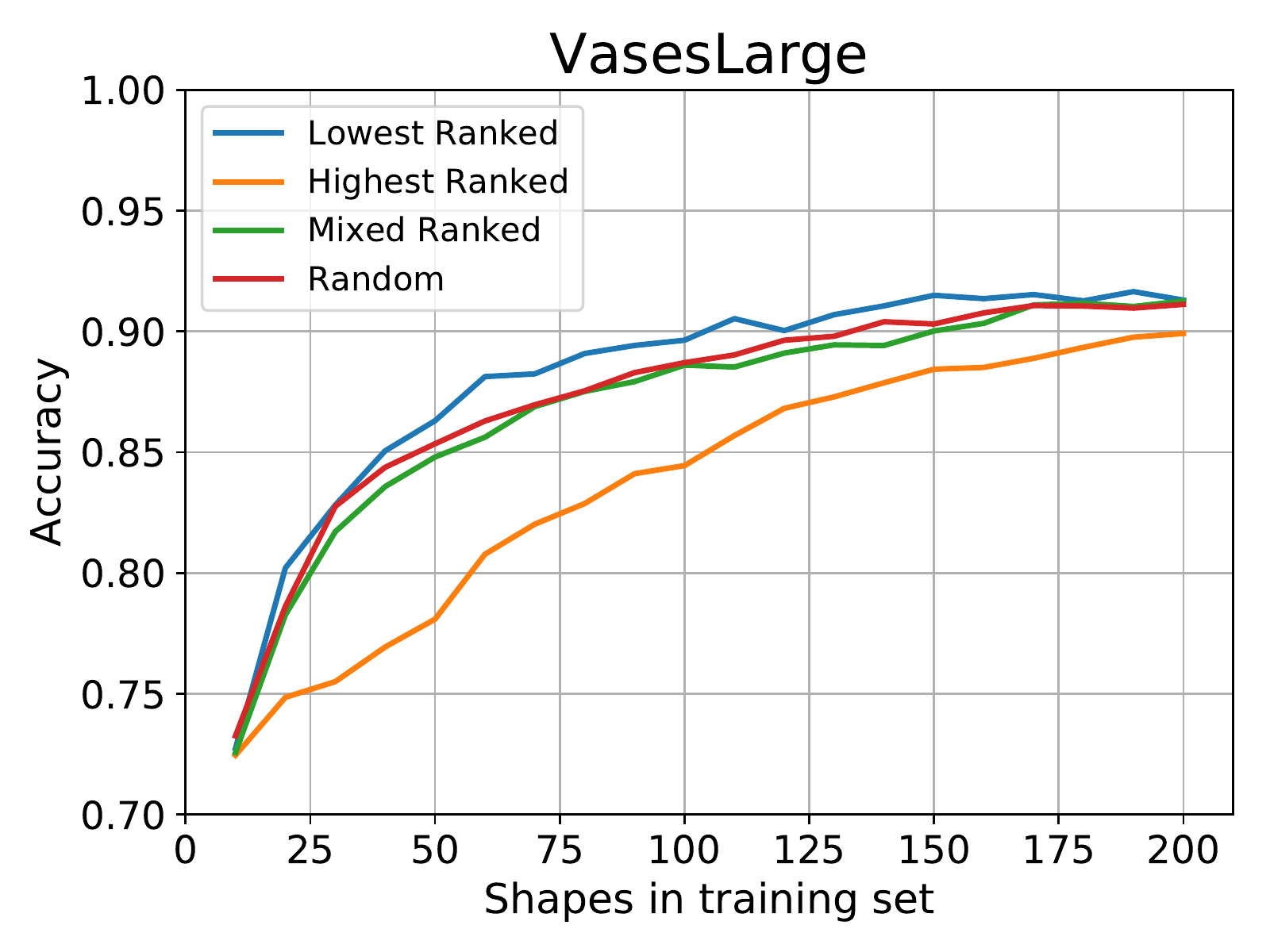}		&
			\includegraphics[width=\size\textwidth]{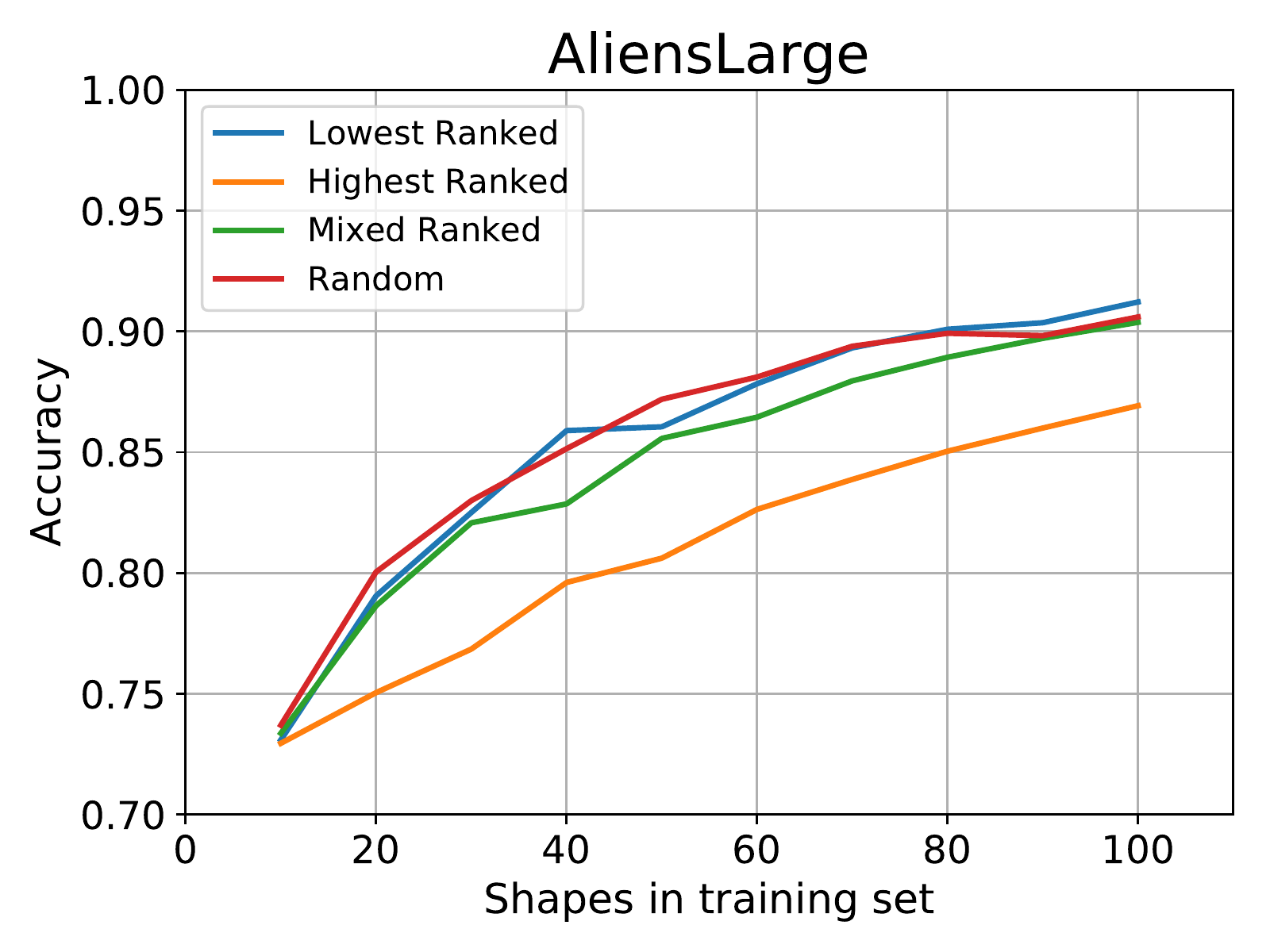}		\\
			\multicolumn{3}{c}{\begin{subfigure}{0.8\textwidth}\caption{Testing Set}\end{subfigure}} 	\\
	\end{tabular}}
	\caption{Results of running our entropy experiments (see text) on the large COSEG datasets \cite{sidi2011}. Each graph shows the average accuracy of all runs for different experiments on different datasets. Shapes are moved from the evaluation set to the training set based on the entropy rank and experiment, while the testing set remains constant throughout the experiment. Each experiment consists of a 5-fold cross validation, where the omitted fold is the testing set and the remaining folds are the training and evaluation sets.}
	\label{fig:entropy_results}
\end{figure*}

\subsection{Entropy Ranking}

Optimally selecting the next set of shapes to refine given the model predictions is key to efficiently labeling a dataset. It might seem logical to select shapes that were predicted well by the model, as these can be worked on quickly. However, it may be more beneficial, overall, to select the shapes the model predicted poorly, as these are the ones that would make the training set more diverse and help the model generalize.

Our entropy ranking algorithm (Section~\ref{sec:table-ordering}) allows for effective ordering of the entire dataset based on the predictions of the model. Using this, we can evaluate the long term effects that selecting high ranking or low ranking shapes has on the model. We devised four experiments to test the entropy ranking, running all experiments on the three large COSEG datasets. Each dataset was split into five equal subsets, and each experiment was ran five times on each dataset, with the results averaged across all runs. For each run, one subset is removed from the dataset and treated as a testing set. The remaining subsets are split into a training set and evaluation set. Each run starts with 10 shapes in the training set, which were selected using our \ac{LFD} \ac{HOG} embedding (Section~\ref{sec:lfd}). The starting training set of each run was fixed across all experiments for a dataset for fairness. Given the starting training set, the model is trained and evaluated on both the evaluation and testing sets, then the evaluation set is ranked according to entropy, and 10 shapes are moved to the training set. The shapes that move depend on the experiment; Lowest Ranked moves the 10 shapes which had the lowest entropy, Highest Ranked moves the 10 shapes which had the highest entropy, Mixed Ranked moves 5 highest and 5 lowest ranked shapes, and Random moves 10 shapes at random. This process is then repeated until 100 shapes remain in the evaluation set. The accuracy of both the evaluation and testing set is recorded each time the model is trained and evaluated, with the results shown in Figure~\ref{fig:entropy_results}. These experiments are intended to mimic use of our program, by using entropy to select shapes to refine. Each time 10 shapes are moved to the training set, we use ground truth labels to train the model, where the user would have labeled them to a ground truth level.

The results shown in Figure~\ref{fig:entropy_results} (a) are the evaluation accuracies for each experiment and each dataset. As is shown, choosing the best ranked shapes will give poor long-term results. This is because the highest ranking shapes are typically similar to shapes already in the training set, so adding these will cause the model to over fit and not generalize. Inversely, choosing the worst ranking shapes gives the best long-term results, as they are typically shapes that have large variation to the training set. Adding these shapes will allow the model to generalize better and prevents over fitting. Finally, the Mixed Ranking and Random results perform similarly, as selecting shapes randomly is likely to contain both high and low ranking shapes, similar to evenly selecting high and low ranking shapes. Additionally, in Figure~\ref{fig:entropy_results} (b), we show the accuracies of the testing set as the training set grows. This shows that all methods except selecting the best ranking methods provide a model which generalize equally to introducing new data to the dataset.

\subsection{Usability and User Study}
\label{sec:user_study}

To test the usability of our system we conducted an in-lab user study which obtained interaction times, clicks and accuracies. We selected 11 participants with good computer skills and provided them with instructions and a demo of how to use the system. 10 of the participants were asked to segment the COSEG \cite{sidi2011} ChairsLarge dataset for approximately 1 hour. We chose this dataset as it contains 400 shapes and has a well defined ground truth segmentation for evaluation of the results. Half of the participants were given the full system, while the other half had the boundary refinement feature disabled. We did this to monitor the usefulness of the feature and its impact on the resulting segmentations. The final participant was asked to segment 6 of the small COSEG datasets, namely, Candelabra, Chairs, Goblets, Guitars, Irons and Lamps. The aim of this experiment was to record times to achieve certain set accuracies for comparisons to previous works.

\begin{figure}[t]
	\centering
	\includegraphics[width=\columnwidth]{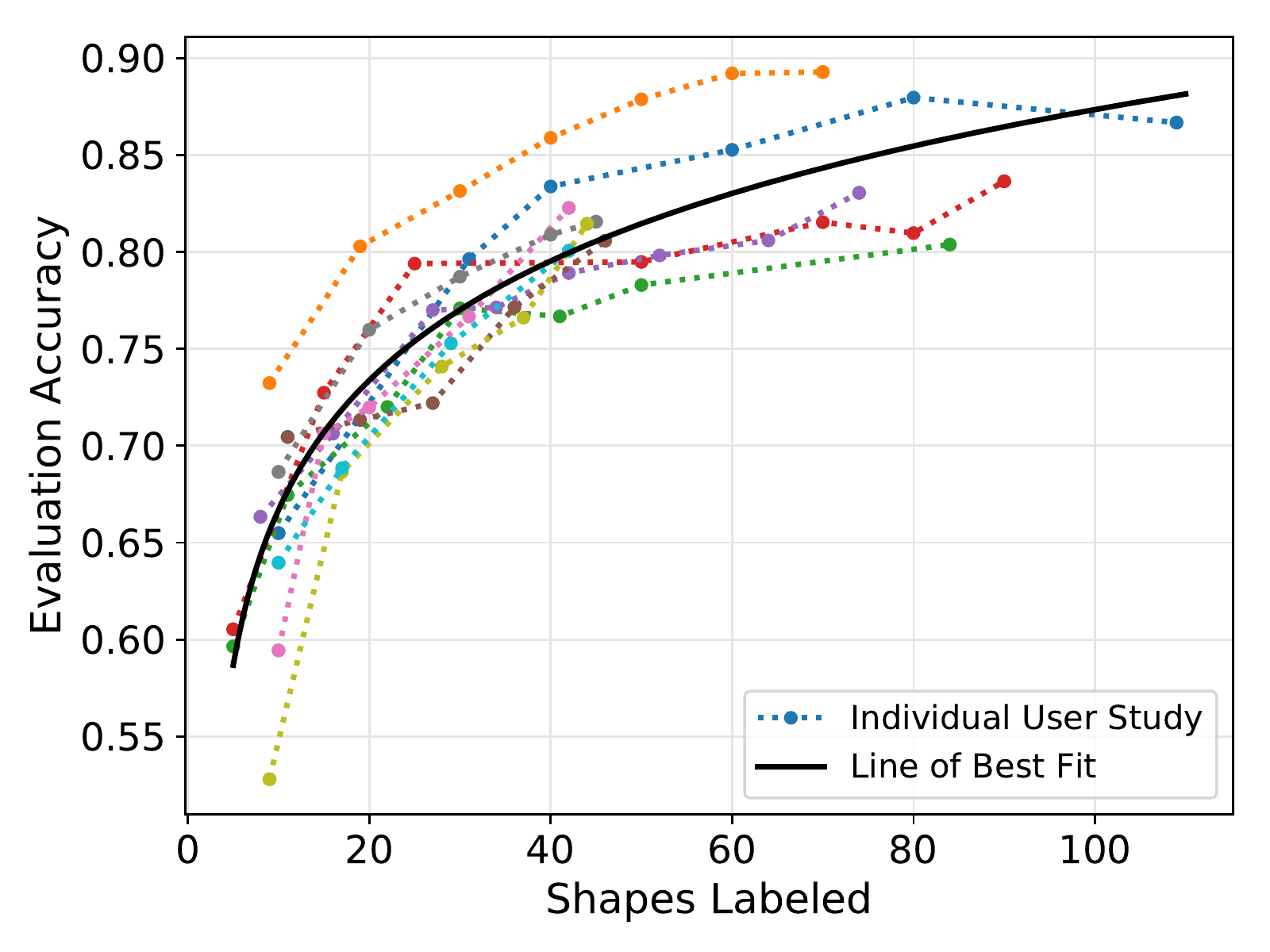}
	\caption{User study results showing accuracy of the evaluation set as the number of shapes in the training set grows. The evaluation set consists of shapes that have not been confirmed by the user. The graph shows that the model generalizes quickly, requiring less work from the user to achieve ground truth accuracy.}
	\label{fig:user_study_accs}
\end{figure}

\begin{figure}[t]
	\centering
	\includegraphics[width=\columnwidth]{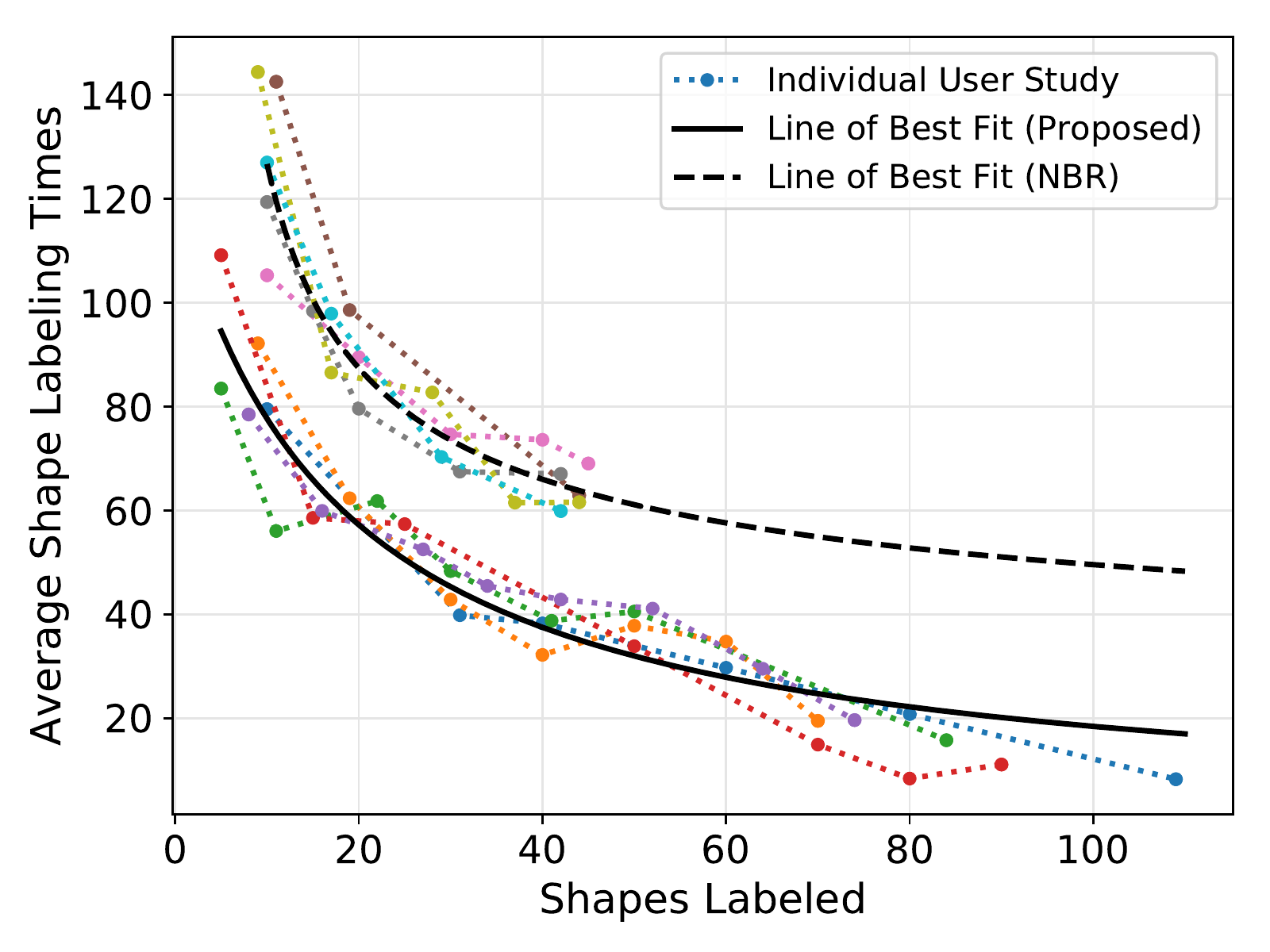}
	
	\includegraphics[width=\columnwidth]{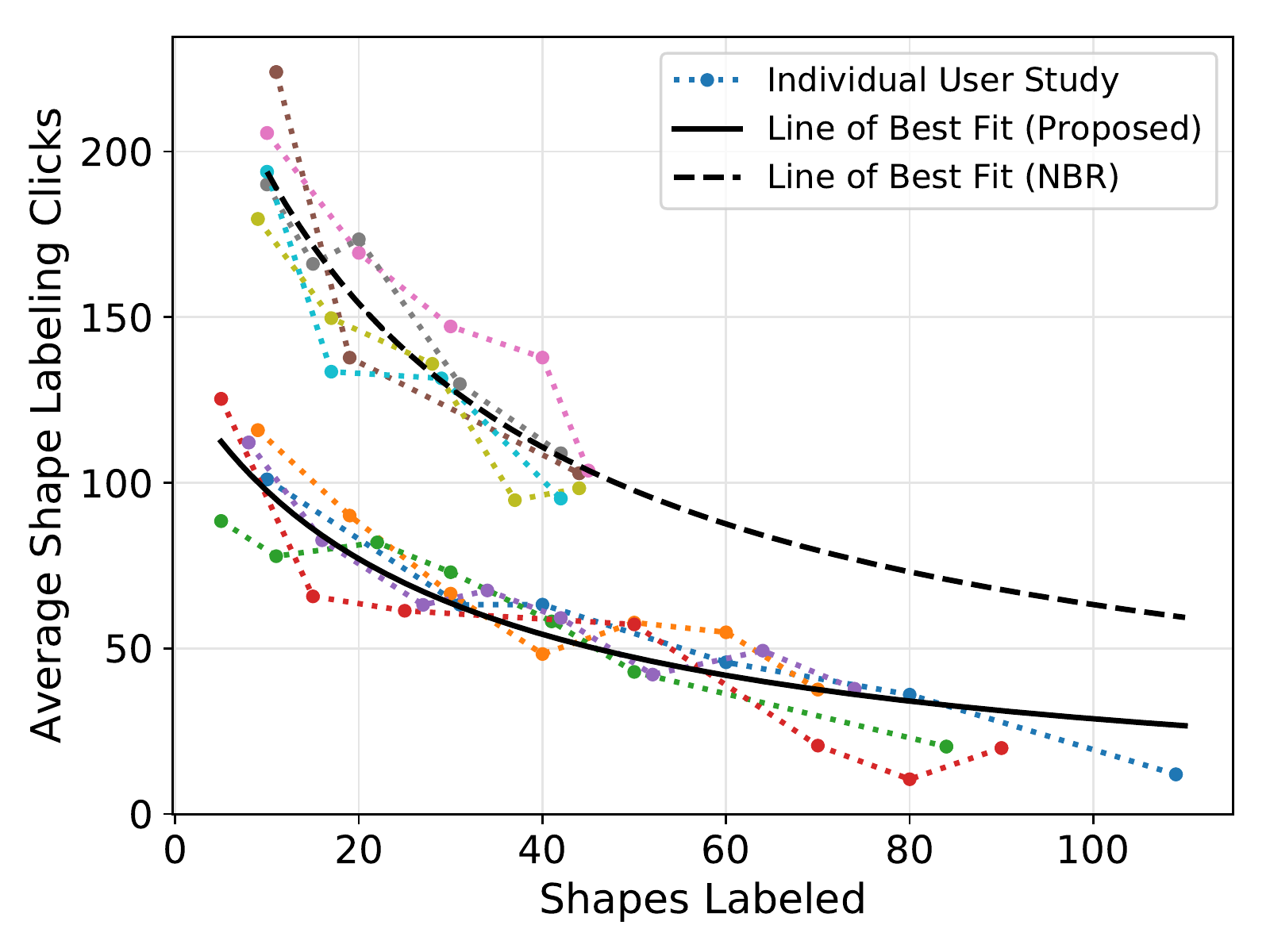}
	\caption{User study results showing interaction time (in seconds) and clicks decreasing as more shapes are added to the training set. The charts also show that users who had the boundary refinement feature disabled (NBR), took longer much longer and required more clicks to achieve the same segmentations, resulting in significantly less shapes segmented in the same time frame}
	\label{fig:user_study_times_clicks}
\end{figure}

The results from the ChairsLarge user studies are shown in Figures~\ref{fig:user_study_accs} and \ref{fig:user_study_times_clicks}. As the results show, the deep learning model quickly gives good performance when evaluated on the remaining shapes, requiring a training set of only 10\% of the dataset to achieve over 80\% accuracy (Figure~\ref{fig:user_study_accs}). Additionally, Figure~\ref{fig:user_study_times_clicks} shows that the required time (a) and interactions (b) to label a shape becomes considerably lower as the model generalizes. Further, there is a significant reduction in labeling times and interactions for the participants who had the boundary refinement feature enabled. This is due to the boundary refinement algorithm providing accurate label outputs.

{\textbf{Comparison to previous work}} Our user study also provided data for comparison to the two previous works, Active Co-analysis (ACA) \cite{wang2012} and Scalable Active Framework (SAF) \cite{yi2016b}. The participant was asked to use the full system and completely label 6 datasets to ground truth level. The times of these experiments were recorded as the dataset accuracy passed certain milestones (i.e. 95\% 98\% etc.) and the results are shown in Table~\ref{table:timing_results}. As shown, our method is comparable to ACA and slightly slower than SAF at achieving a 95\% accuracy. However, as the purpose of this work is providing a tool for efficient ground truth generation, 95\% accuracy is not a good enough segmentation. We show this in Figure~\ref{fig:timing_comparison}, where the average accuracy of the set may be 95\%, but certain individual shapes show very poor segmentations. For these reasons we also provide timings to achieve higher set accuracies, including a ground truth level (100\%). Our method can achieve this level of segmentation as it not only asks the user to verify the results but correct any mistakes with a refinement stage. 

\begin{table}[t]
	\newcommand{\tbf}{\textbf}
	{\resizebox{\columnwidth}{!}{    
			\begin{tabular}{@{}L{1.8cm} @{}C{1cm} @{}C{1cm} @{}C{1cm} @{}C{1cm} @{}C{1cm} @{}C{1cm}@{}}	
				\toprule
				& \tbf{ACA}			& \tbf{SAF}			& \multicolumn{4}{c}{\tbf{Proposed}}										\\ \cmidrule[\heavyrulewidth]{4-7}
				& \tbf{95\%}		& \tbf{95\%}		& \tbf{95\%}		& \tbf{98\%}		& \tbf{99\%}		& \tbf{100\%}	\\ \toprule
				\tbf{Candelabra}	& 7.00 				& 1.40	     		& 5.56				& 6.34	     		& 7.47				& 8.17			\\ \midrule
				\tbf{Chairs}   		& 10.50*			& 0.90*     		& 1.30 	    		& 1.99				& 2.21				& 2.88			\\ \midrule
				\tbf{Goblets}  		& 1.20*				& 0.70*     		& 1.01 	    		& 1.51				& 1.51				& 1.54			\\ \midrule
				\tbf{Guitars}  		& 1.80*				& 1.90*     		& 2.37 	    		& 5.58				& 6.35				& 9.19			\\ \midrule
				\tbf{Irons}    		& 7.60*				& 7.20*     		& 2.63 	    		& 3.27				& 3.27				& 3.58			\\ \midrule
				\tbf{Lamps}    		& 0.60*				& 2.30*     		& 3.14  	   		& 3.82				& 4.54				& 5.26			\\ \bottomrule
	\end{tabular}}}
	\captionof{table}{Comparison of user interaction times for achieving certain dataset accuracies. We compare our method the two previous work, ACA \cite{wang2012} and SAF \cite{yi2016b} (* denotes estimated times, see original papers). While our method performs similar to ACA and slightly worse than SAF, we strive for high quality segmentations and good boundaries. Furthermore, reporting 95\% accuracy is not ground truth level, so we also report times to achieve accuracies up to ground truth level.}
	\label{table:timing_results}
\end{table}

{\textbf{Computation Time}} Our estimated computation times are based on using our system to label the COSEG ChairsLarge dataset. Our pre-processing stage consists of manifold checking (<0.1s per shape) and feature extraction (\textasciitilde40s per shape). Then, our deep learning model takes \textasciitilde90s to train (This time is fixed due to our training scheme) and <0.25s per shape to evaluate and refine. In future we would refine the training and evaluation process to run concurrently with user interactions, greatly reducing the processing time. This scales linearly with the size of the dataset. All timings are reported using a 4-core 4GHz Intel Core i7, 32GB of RAM and a Nvidia GTX 1080Ti with 11GB of VRAM.

\begin{figure}[t]
	\centering
	\begin{subfigure}[t]{\columnwidth}
		\includegraphics[width=\textwidth]{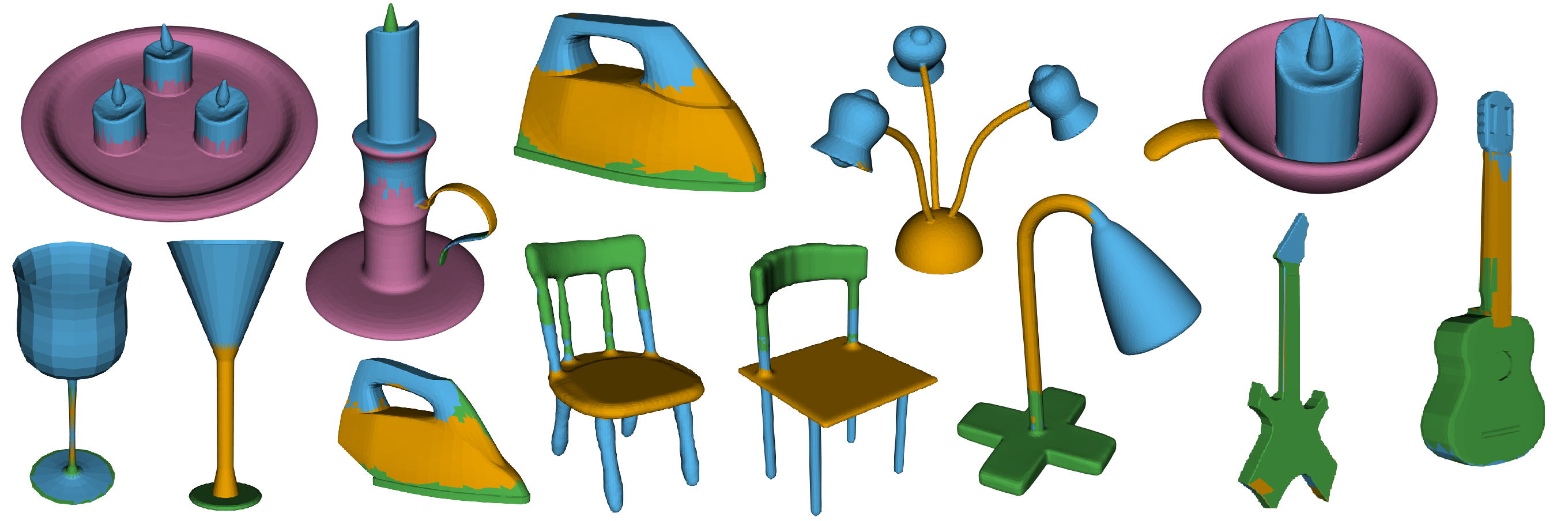}
		\caption{95\% accuracy}
	\end{subfigure}
	
	\begin{subfigure}[t]{\columnwidth}
		\includegraphics[width=\textwidth]{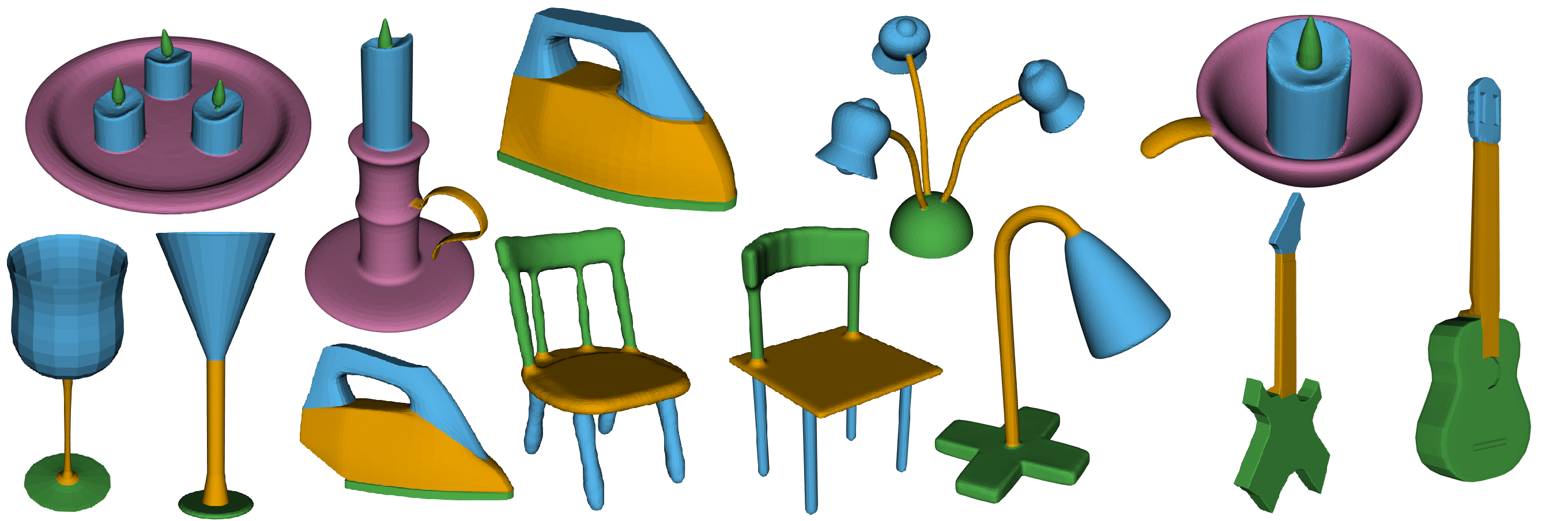}
		\caption{Ground truth accuracy}
	\end{subfigure}
	\caption{Visual comparison of segmentation results from sets labeled to 95\% accuracy (a) and to ground truth level (b). This shows that a set labeled to 95\% accuracy still requires significant work to complete, therefore times to achieve this accuracy are less meaningful}
	\label{fig:timing_comparison}
\end{figure}

\subsection{ShapeNet Labeling}

To evaluate the usability of our system on very large datasets, we use ShapeNet \cite{chang2015} datasets. However, as many of the shapes are non-manifold we have re-meshed several of the datasets for these experiments (See Section~\ref{sec:shapenet}). To ensure our system is usable with very large datasets we chose the Airplane and Guitar datasets, and included 4 smaller datasets for a more thorough evaluation of the re-meshing quality.

\begin{table}[t]
	\newcommand{\tbf}{\textbf}
	{\resizebox{\columnwidth}{!}{    
			\begin{tabular}{@{}L{1.8cm} 	@{}C{1.2cm} 	@{}C{1.2cm} 	@{}C{1.2cm}		@{}C{1.2cm} 			@{}C{1.2cm}		@{}C{1.2cm}@{}} 
				\toprule
				&				&				& \multicolumn{2}{c}{\tbf{Proposed}}	& \multicolumn{2}{c}{\tbf{SAF}}		\\
				& \tbf{N} 		& \tbf{NR}		& \tbf{L}		& \tbf{T (Hrs)}			& \tbf{L}		& \tbf{T (Hrs)}		\\ \toprule
				\tbf{Airplane}	& 4027	 		& 4009	    	& 6		  		& 24.1*					& 4				& 22.6*				\\ \midrule
				\tbf{Bag}   	& 83			& 75	    	& 2		  		& 0.4					& 2				& 0.2* 				\\ \midrule
				\tbf{Cap}		& 56			& 55	    	& 2		  		& 0.3					& 2				& 0.2* 				\\ \midrule
				\tbf{Earphone} 	& 73			& 60	    	& 4		  		& 0.3					& 2				& 0.2* 				\\ \midrule
				\tbf{Guitar}  	& 793			& 794	    	& 3		  		& 3.0*	 				& 3				& 2.8*				\\ \midrule
				\tbf{Knife}    	& 426			& 420	    	& 2		  		& 1.7*	 				& 2				& 1.5*				\\ \bottomrule
	\end{tabular}}}
	\captionof{table}{Re-mesh and labeling statistics for ShapeNet datasets. For each dataset we report, number of shapes (N), number of successfully re-meshed shapes (NR), number of labels (L) and the (user interaction) time to label the dataset in hours (T). Any times shown with (*) are estimated based on labeling the dataset for 1 hour}
	\label{table:shape_net}
\end{table}

\begin{figure}[t]
	\centering
	\begin{subfigure}[t]{\columnwidth}
		\includegraphics[width=\textwidth]{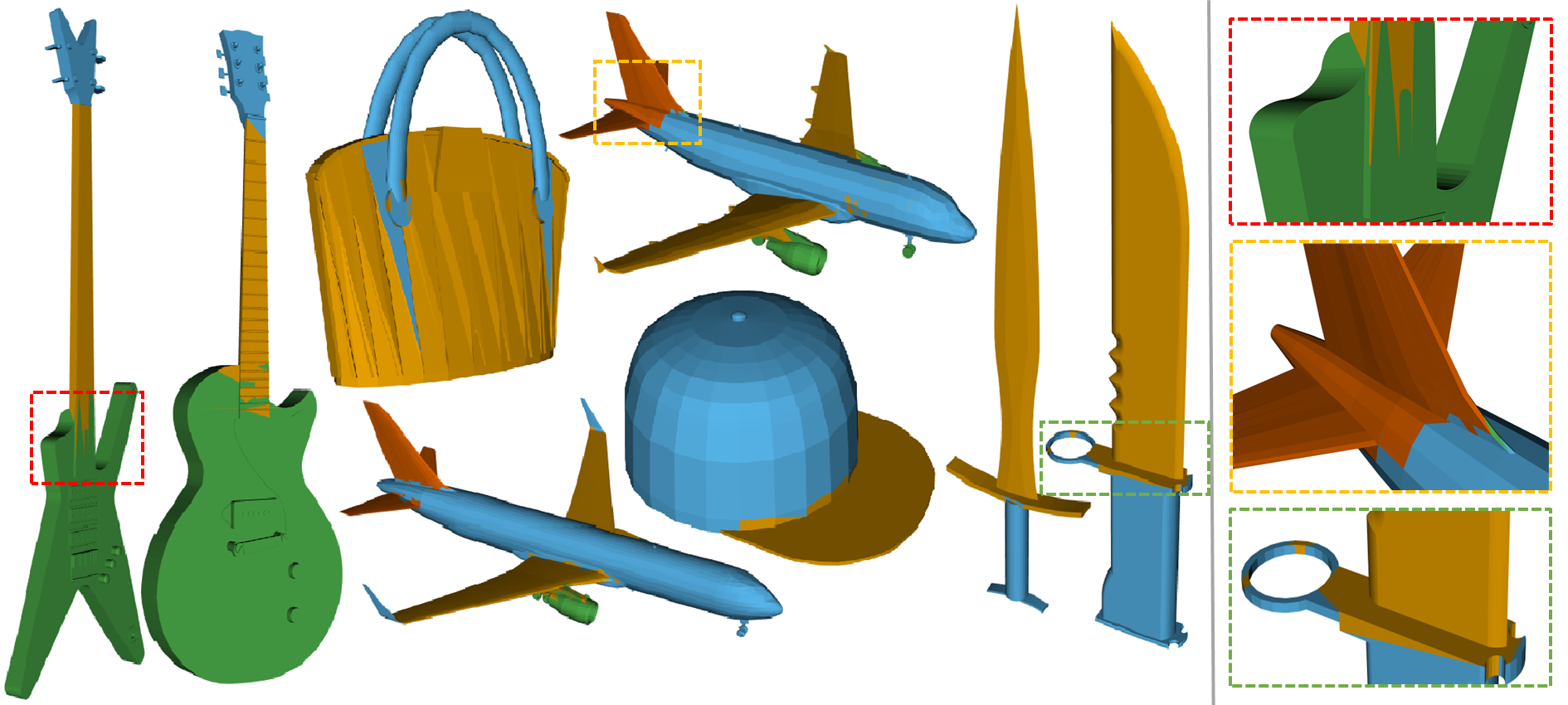}
		\caption{SAF \cite{yi2016b}}
	\end{subfigure}
	\begin{subfigure}[t]{\columnwidth}
		\includegraphics[width=\textwidth]{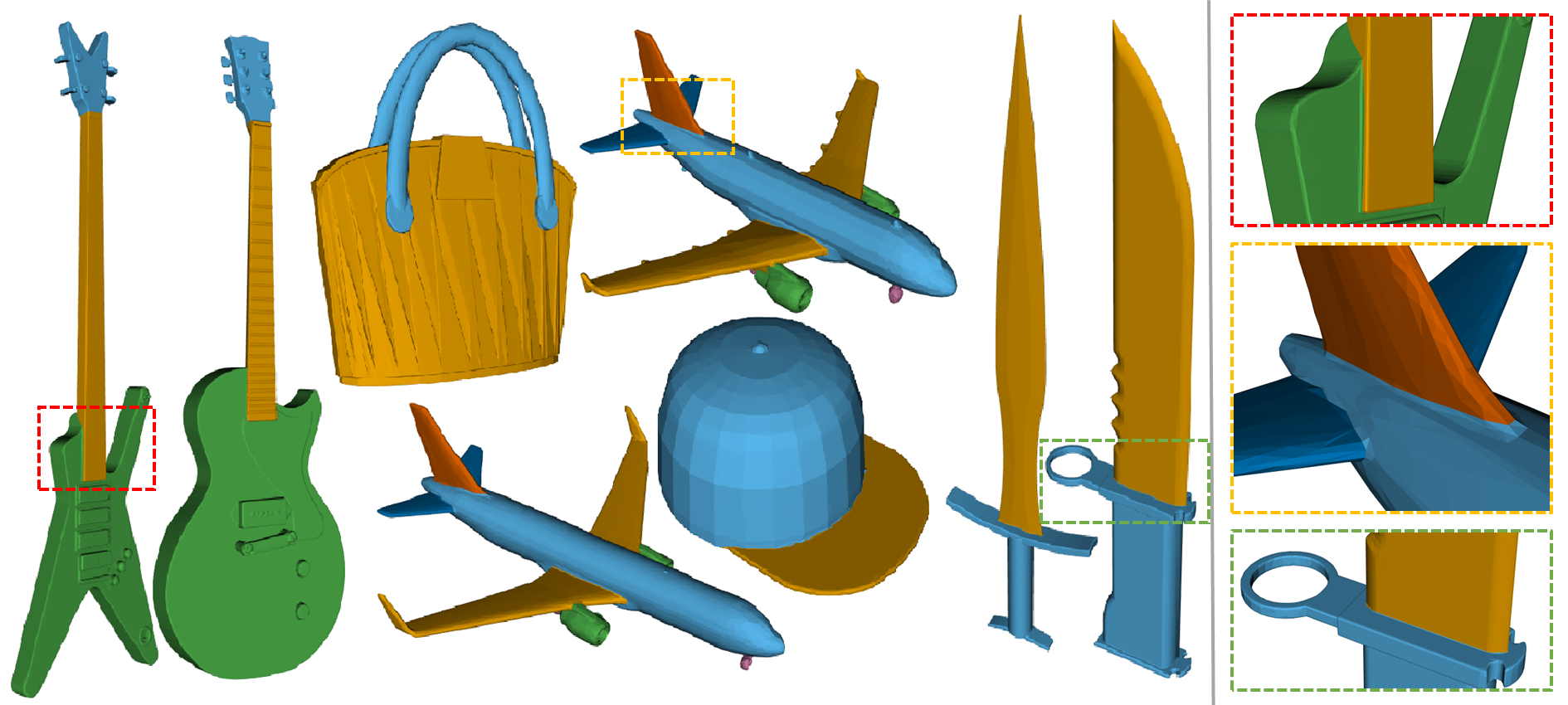}
		\caption{Proposed}
	\end{subfigure}
	\caption{Visual comparison of segmentation results from ShapeNet datasets. We compare provided segmentations from \cite{yi2016b} (a) to segmentations generated by our proposed framework (b). Highlighted regions are shown on the right in a zoomed view. As is shown, our method can provide much more accurate segment boundaries with our refinement step.}
	\label{fig:shape_net_comparison}
\end{figure}

Our experiments are formulated similar to our User Studies (Section~\ref{sec:user_study}), where our system was used for up to one hour of user time for each dataset. Table~\ref{table:shape_net} shows the timings achieved when labeling the 6 ShapeNet datasets (times shown with (*) are estimated based on 1 hour of user time). The table also shows the number of shapes that were successfully re-meshed and the number of labels the dataset had. The results show that our method is slightly slower than SAF \cite{yi2016b}, however we emphasize high quality segmentation, so additionally we compare the quality of the output segmentation of each system. As Figure~\ref{fig:shape_net_comparison} shows, there is a significant difference in segment boundary quality between the two methods. Our method maintains good quality boundaries, while the segmentation from SAF is poor in some regions. While this poor segmentation could be due to point cloud resolution or label projection, Figure~\ref{fig:shape_net_comparison} shows that there are also many cases of poor labeling on the point cloud. Additionally, we also label the Airplane and Earphone datasets with an increased number of segments (6 instead of 4 for Airplane, 4 instead of 3 for Earphone). For SAF to achieve this number of segments, much more user time would be required (each new label requires a full pass of the dataset to be processed). Therefore, our time is likely comparable or faster to what SAF would achieve with the same segmentations. This shows that as the number of segments increase, our system can outperform SAF in both quality and efficiency.


\section{Conclusions}
\label{sec:conclusion}

In this work, we have shown and efficient and accurate active learning framework driven by a fast and effective deep learning model. The motivation behind this work was providing high quality shape segmentations for very large datasets in an efficient way. To achieve this we combine three core systems; a deep learning model, effective shape ordering and selection, and accurate refinement tools. These three systems combine to create an iterative interactive framework, which becomes more effective over time.

We have shown that our framework is not only more accurate than the current state-of-the-art, but also more efficient for datasets with large amounts of distinct segments. We also demonstrate that our system can scale with massive datasets, allowing for quick and meaningful segmentation of thousands of shapes.


\begin{figure}[t]
	\centering
	\begin{subfigure}[t]{\columnwidth}
		\includegraphics[width=\textwidth]{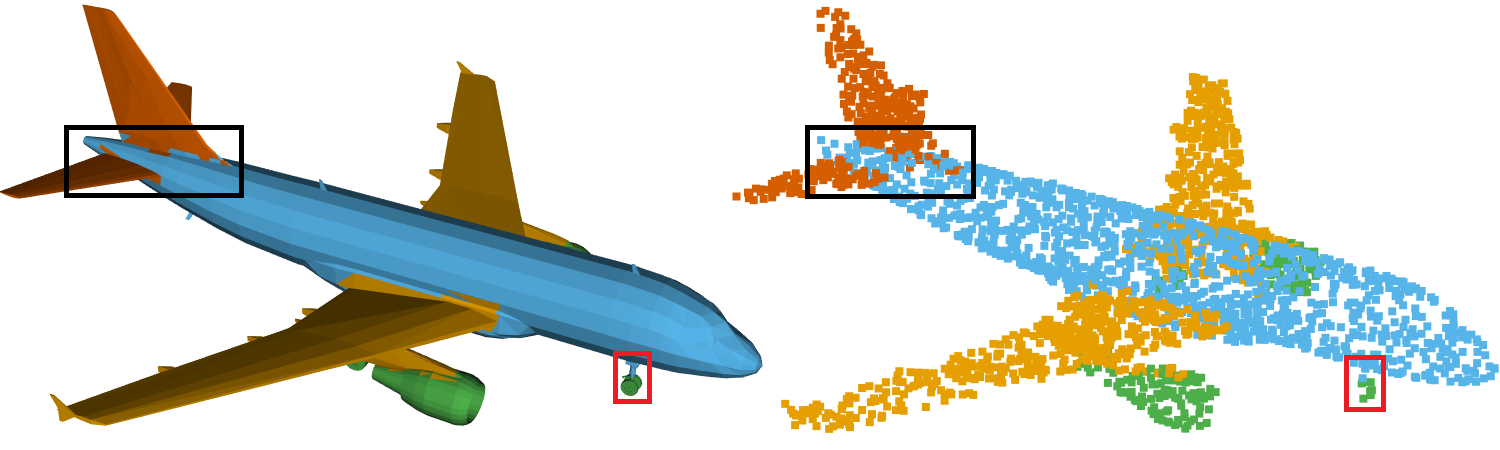}
	\end{subfigure}
	
	\begin{subfigure}[t]{\columnwidth}
		\includegraphics[width=\textwidth]{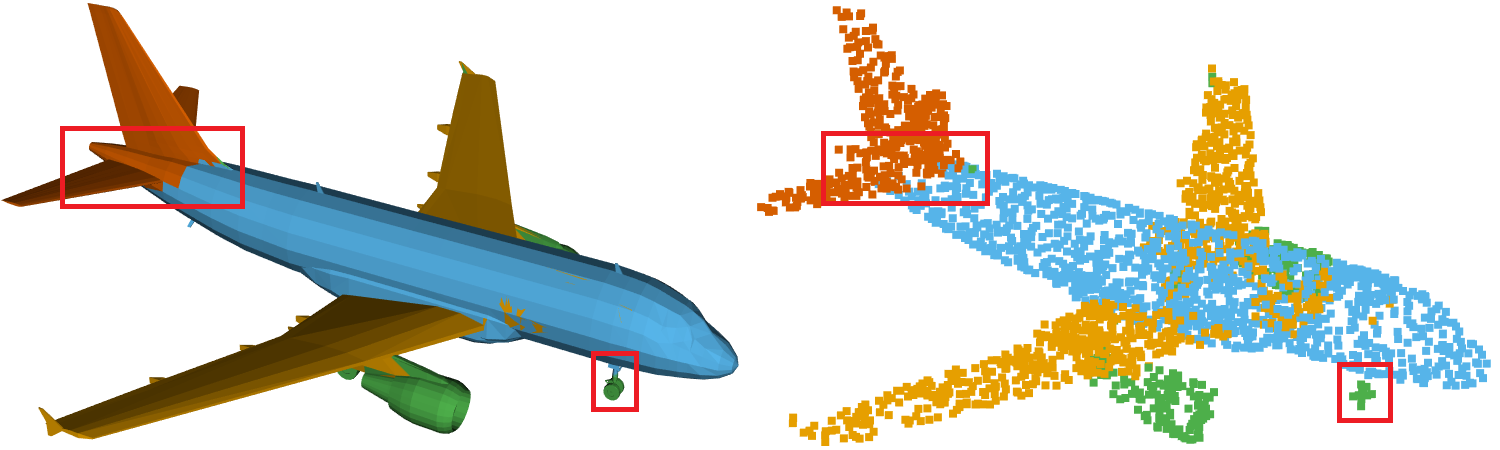}
	\end{subfigure}
	
	\begin{subfigure}[t]{.48\columnwidth}
		\includegraphics[width=\textwidth]{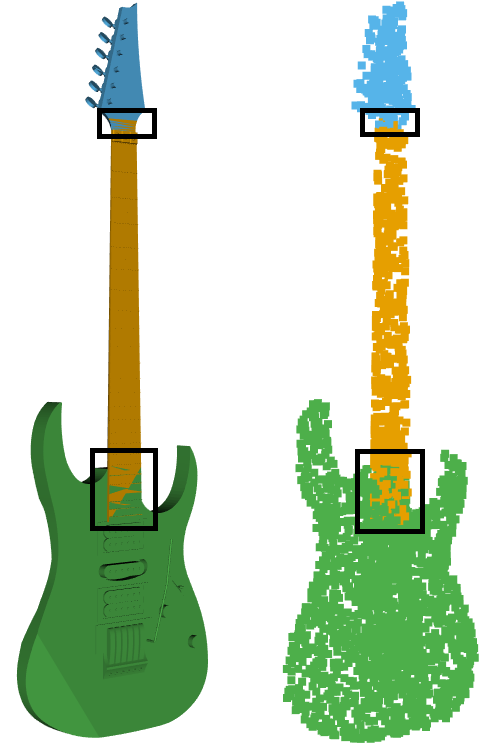}
	\end{subfigure}
	\begin{subfigure}[t]{.48\columnwidth}
		\includegraphics[width=\textwidth]{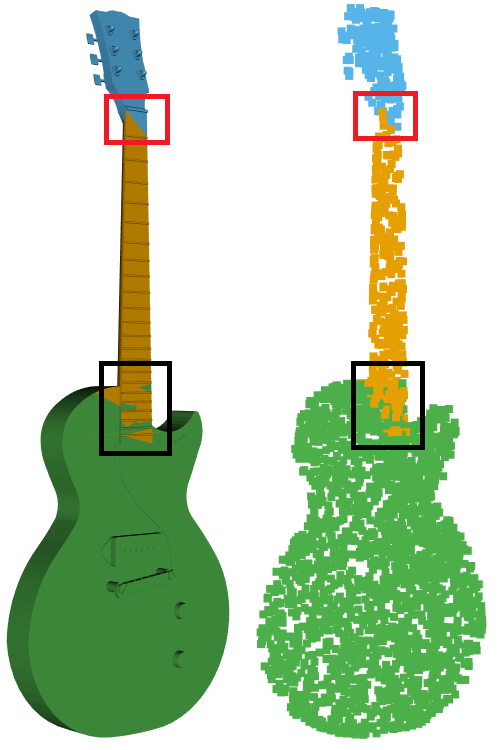}
	\end{subfigure}
	\caption{Comparison between provided ShapeNet labels from SAF \cite{yi2016b}, when displayed on point clouds or projected onto the original mesh. While there are cases where point cloud resolution impacts the projection (black), there are also many incorrectly labeled sections (red)}
	\label{fig:saf_comparison}
\end{figure}

\subsection{Limitations and Future Work}

A trade-off we made with our pipeline was full dataset evaluation. This enables us to provide a powerful shape ordering tool at the cost of processing time. While current dataset sizes do not pose a major time delay for evaluation, as datasets continue to grow, it could soon be an issue. There are several ways we could resolve this while still maintaining effective shape ordering. One way, would be to only evaluating on a subset of the data. The selection of the subset would then be key to maintaining effective shape ordering. A solution would be using global shape descriptors to select shapes both similar and dissimilar to shapes in the training set. However, the size of the required subset would still need to be large so that the user has enough diversity when selecting shapes to refine. A better solution would be to train and evaluate in the background. This solution would minimize any user down time while still providing an always up-to-date table. As shapes are confirmed they can be trained on quickly, then shapes can be evaluated (according to shape similarity) and the table can be dynamically updated.

Another trade-off we made was requiring manifold shapes. While re-meshing software is available, and we show a working method in this work, this is still not ideal. The main reasons we require manifold shapes are feature extraction and user painting. Other works have converted the shapes to point clouds, and while this fixes any topology issues, there is still information loss in this process. Another solution would be to introduce artificial edges in the shapes to join the components. While this doesn't solve all the problems in ShapeNet datasets (such as zero-thickness parts and low resolution), it would allow for features to be extracted and painting between parts. 

\bibliographystyle{ACM-Reference-Format}
\bibliography{references}

\end{document}